\documentclass[review]{elsarticle}

\usepackage{lineno,hyperref}
\modulolinenumbers[5]
\usepackage[T1]{fontenc}
\usepackage{graphicx}
\usepackage{amssymb,amsmath,amsthm}
\usepackage{mathtools}
\usepackage{booktabs}	
\usepackage{float}	
\usepackage{subcaption} 
\usepackage{enumitem}
\usepackage{bm}
\usepackage{color,soul} 
\usepackage{array}
\usepackage[margin=0.9in]{geometry} 
\usepackage{siunitx} 
\usepackage{cancel} 
\usepackage{comment} 
\biboptions{numbers,sort&compress}
\usepackage[capitalize]{cleveref}
\usepackage[toc,page]{appendix}
\usepackage{tikz}
\usepackage{bm}
\usetikzlibrary{arrows.meta}
\usepackage{algorithm}
\usepackage{algpseudocode}
\usepackage{multirow}
\usepackage{siunitx}
\usepackage[table]{xcolor}
\usepackage[draft,inline,nomargin,index]{fixme}
\fxsetup{theme=color,mode=multiuser}
\FXRegisterAuthor{rt}{arn}{\color{blue} RT}
\FXRegisterAuthor{sn}{asn}{\color{red} SN}

\newcolumntype{d}[1]{D{.}{.}{#1}}
\newcommand{\matr}[1]{\mathbf{{#1}}}
\newcommand{\ve}[1]{\bm{{#1}}}

\newcommand{\brc}[1]{\left(#1\right)}
\newcommand{\cbk}[1]{\left\{#1\right\}}
\newcommand{\sbk}[1]{\left[#1\right]} 

\DeclareMathOperator{\zeros}{\matr{0}}

\newcommand{\pd}{\partial}

\begin{document}

\begin{frontmatter}

\title{Machine learning assisted state prediction of misspecified linear dynamical system via modal reduction}

\author[label1]{Rohan Vittal Thorat}
\address[label1]{Department of Applied Mechanics, Indian Institute of Technology Delhi, New Delhi, 110016, India}
\ead{rohan.thorat@am.iitd.ac.in}
\author[label1]{Rajdip Nayek}
\ead{rajdipn@am.iitd.ac.in}

\cortext[cor1]{Corresponding author}

\begin{abstract}
\sl

\noindent Accurate prediction of structural dynamics is imperative for preserving digital twin fidelity throughout operational lifetimes. Parametric models with fixed nominal parameters often omit critical physical effects due to simplifications in geometry, material behavior, damping, or boundary conditions, resulting in \textit{model form errors} (MFEs) that impair predictive accuracy. This work introduces a comprehensive framework for MFE estimation and correction in high-dimensional finite element (FE) based structural dynamical systems. The Gaussian Process Latent Force Model (GPLFM) represents discrepancies non-parametrically in the reduced modal domain, allowing a flexible data-driven characterization of unmodeled dynamics. A linear Bayesian filtering approach jointly estimates system states and discrepancies, incorporating epistemic and aleatoric uncertainties. To ensure computational tractability, the FE system is projected onto a reduced modal basis, and a mesh-invariant neural network maps modal states to discrepancy estimates, permitting model rectification across different FE discretizations without retraining. Validation is undertaken across five MFE scenarios—including incorrect beam theory, damping misspecification, misspecified boundary condition, unmodeled material nonlinearity, and local damage —demonstrating the surrogate model’s substantial reduction of displacement and rotation prediction errors under unseen excitations. The proposed methodology offers a potential means to uphold digital twin accuracy amid inherent modeling uncertainties.

\end{abstract}


\end{frontmatter}

\section{Introduction}
The reliable simulation of structural dynamical systems is central to engineering analysis, design, and decision-making. In practice, high-fidelity models are often impractical due to limited information, computational constraints, or simplifying assumptions in geometry, boundary conditions, damping mechanisms, and material constitutive laws. These idealizations lead to model form errors (MFEs)---systematic discrepancies between the predicted and actual system responses---which, if unaccounted for, can significantly degrade predictive accuracy. This challenge is especially critical in the context of digital twins, where model predictions directly inform monitoring and decision-making.

Digital twins of structural systems integrate computational models with real-time or historical measurement data to enable continuous prediction, monitoring, and decision making \cite{vanderhorn2021digital,tao2018digital}. Yet for a digital twin to remain reliable throughout its operational life, the underlying simulation model must reproduce system dynamics with high fidelity. In practice, however, the nominal computational models, although often furnished with reasonably accurate parameter values, inevitably omit certain physical effects due to simplifying assumptions, sometimes intentionally to reduce computational cost and other times unintentionally due to lack of expert knowledge. These missing terms introduce MFEs that can compromise predictive reliability when extrapolating beyond the calibration regime.

A common strategy in digital twin updating is to recalibrate model parameters so that simulation outputs match observed data \cite{sung2024review,wong2017frequentist,xie2021bayesian}. Such parameter tuning can improve short-term agreement, but it leaves a fundamental gap: MFEs that arise from neglected physics or unmodeled couplings are not parameterizable within the chosen model form. Attempting to absorb these effects through model parameter calibration risks biasing otherwise well-characterized parameters, particularly when the number of tunable parameters is far fewer than the model dimension. Thus, while model parameter calibration remains useful when uncertainty lies primarily in parameters, it is insufficient when the governing equations themselves are misspecified.
Our focus, therefore, is on \textit{estimating and compensating for such missing modeling terms directly}, without altering the known parameters of the nominal model. This strategy preserves the structural integrity of the nominal computational model while enabling data-driven rectification of its deficiencies, which can also address model parameter-based errors as part of the inferred discrepancy, ensuring that predictive accuracy is not sacrificed even when model parameter errors coexist with unmodeled physics.


Over the past two decades, Gaussian Process (GP) modeling has emerged as a powerful probabilistic framework for augmenting imperfect computational models. GP-based approaches for handling model error can be broadly classified into three categories. The first places a GP discrepancy term in the measurement equation, as in the Kennedy–O'Hagan (KOH) framework \cite{kennedy2001bayesian}, where model outputs are statistically calibrated against data. This is suitable when the nominal simulator is a black-box and governing equations are inaccessible; however, this approach only corrects outputs and does not amend deficiencies at the dynamics level \cite{maupin2020model,higdon2004combining,higdon2008computer,brynjarsdottir2014,arendt2012quantification}.

The second category embeds the GP within the process equations, thereby acting directly on the dynamics rather than the measurements. A prominent example is the class of latent force models (LFMs) \cite{alvarez2013linear}, in which a GP prior is imposed on unobserved forcing terms in ODE or PDE systems. By correcting the governing equations themselves, LFMs offer greater generalizability than KOH-type corrections, though they require intrusive access to the governing equations. LFMs have been applied in structural dynamics to represent unknown inputs and/or discrepancies \cite{nayek2019gaussian,2022RogersNonlinearIdentification, kashyap2024gaussian, bisaillon2022combined, garg2022physics,lourens5295147bias}, but these applications have been restricted to small-scale systems, with structural degrees of freedom (DOFs) typically less than ten. The computational difficulty of scaling LFM inference to high-dimensional FE systems (since each DOF typically incurs its own latent force) has left open the challenge of extending such methods to realistic structures. Moreover, beyond inference, only Garg et al.~\cite{garg2022physics} attempted to use the inferred discrepancies for forward prediction, and even then, the application was limited to small-scale MDOF systems. Extending such rectification to high-dimensional FE-based models, therefore, remains relatively unexplored.


A third, more recent, line of work places GP priors directly on the solution field, enforcing physical constraints through covariance structure or weak forms of the governing equations \cite{raissi2019physics,pfortner2022physics,hennig2015probabilistic,cockayne2016probabilistic,oates2019modern,tronarp2019probabilistic,long2022autoip}. While this approach is attractive for linear problems, where closed-form inference can be derived, it becomes computationally demanding in nonlinear settings.

Among these alternatives, LFMs are particularly relevant for our objective of inferring model discrepancy and rectifying imperfect dynamics, since they operate directly at the dynamics level (and are not restricted to just the measurement level). Yet their extension to high-dimensional structural dynamics remains underdeveloped.

In this work, we extend the LFM framework for inferring model discrepancy \cite{kashyap2024gaussian} to high-dimensional linear structural dynamical systems. Our approach has two key elements:  
(a) the high-dimensional nominal linear system is reduced via modal decomposition, and GPLFMs are applied to the reduced-order modal equations to infer latent modal discrepancy forces, so that the number of latent variables scales with the number of linear elastic modes rather than the number of DOFs; and  
(b) the inferred discrepancies are mapped to the corresponding modal states through a neural network surrogate, providing a \textit{discretization-mesh-invariant correction} that can be embedded back into the nominal computational model for predictive rectification under unseen excitations.  

Although, in principle, black-box machine learning models such as neural networks or neural ODEs could also be used to learn discrepancies, stationary GPs are particularly advantageous in this setting. They provide flexible representations of dynamical signals with only a few hyperparameters, and crucially, admit equivalent linear stochastic differential equation representations. This allows smooth integration with reduced-order linear dynamics and efficient Kalman filtering for joint state–discrepancy inference. These properties make GPLFMs especially well-suited for large-scale linear structural dynamics. This design enables scalable and mesh-invariant rectification of linear structural dynamical models, extending LFM-based error inference from lumped or low-dimensional systems to realistic FE-based digital twins.

The primary contributions of this study are:
\begin{itemize}
    \item Non-parametric black-box modeling of MFEs in the modal domain via GPLFM.

    \item Mesh-invariant model correction via a neural network mapping from reduced-order modal states to MFEs, transferable across FE meshes of varying discretizations.

    \item Physics-augmented simulation rectification by embedding a learned surrogate into the nominal model, significantly improving response predictions under unseen loading conditions.

    \item Comprehensive numerical validation through diverse case studies, including incorrect governing equations, damping misspecification, boundary condition misspecification, unmodeled material nonlinearity, and local damage.
\end{itemize}

The remainder of the paper is organized as follows: \cref{sec:problemstatement} formalizes the dynamical system of interest and states the estimation and prediction objectives in a precise manner. \cref{sec:methodology} formulates the reduced-order modal representation, GP-based discrepancy model, details the augmented state-space framework and Kalman filter-based inference, and presents the neural surrogate training and mesh-invariant correction strategy. \cref{sec:numerical_examples} reports numerical studies and performance evaluation. \cref{sec:discussion} covers discussion and \cref{sec:conclusion} concludes with future research directions.

\section{Problem Statement} \label{sec:problemstatement}

Consider a high-dimensional structural dynamical system governed by the following equations of motion obtained from the finite element (FE) discretization of a continuous structure:
\begin{equation}
    \matr{M} \ddot{\bar{\ve{u}}}(t) + \matr{C} \dot{\bar{\ve{u}}}(t) + \matr{K} \bar{\ve{u}}(t) + \ve{g}\brc{\bar{\ve{u}}(t), \dot{\bar{\ve{u}}}(t)} = \ve{f}(t), \quad 0 \le t \le T
    \label{eq:trueFE}
\end{equation}
where $t$ is time in the range on interest $[0, T]$ and $\bar{\ve{u}}(t) \in \mathbb{R}^N$ is the unknown true displacement vector of $N$ degrees of freedom (DOFs). The system is assumed to be constrained such that no rigid-body modes exist. The vectors $\dot{\bar{\ve{u}}}(t) = \frac{d\bar{\ve{u}}(t)}{dt}$ and $\ddot{\bar{\ve{u}}}(t) = \frac{d^2\bar{\ve{u}}(t)}{dt^2}$ denote velocity and acceleration, respectively. The matrices $\matr{M}, \matr{C}, \matr{K} \in \mathbb{R}^{N\times N}$ represent the mass, damping, and stiffness operators, and are assumed to be known, symmetric, and positive definite. The term $\ve{f}(t) \in \mathbb{R}^N$ denotes the external forcing vector. Without any loss of generality, one may assume zero initial conditions.

The vector-valued function $\ve{g}\brc{\bar{\ve{u}}(t), \dot{\bar{\ve{u}}}(t)}$ represents \emph{additional restoring forces} beyond the nominal linear contributions. This term is a placeholder for physical effects that enrich the system description, such as nonlinear material constitutive behavior, geometric nonlinearities, higher-order theory effects (e.g., shear deformation or rotary inertia in Timoshenko beams), non-classical damping mechanisms, contributions from unmodeled structural components (e.g., joints or springs), and local stiffness variations associated with damage. When included, $\ve{g}$ yields a more faithful representation of the true dynamics. Its omission, however, results in systematic discrepancies between the nominal and true responses, which we identify as \textit{model-form errors} (MFEs).

We assume the knowledge of the nominal parameters $\brc{\matr{M}, \matr{C}, \matr{K}}$ and the initial state of the system at $t=0$.

With this setup, the objectives are twofold:
\begin{itemize}
    \item \textbf{Estimation:} To estimate the system response trajectory $\bar{\ve{u}}(t)$ of the true system \cref{eq:trueFE} for a given excitation $\ve{f}(t)$ by reconstructing the missing contribution $\ve{g}\brc{\bar{\ve{u}}(t),\dot{\bar{\ve{u}}}(t)}$ from measurements of system responses in the form of displacement time series, denoted by the dataset $\mathcal{D} = \cbk{(t_k,\ve{y}_k)}_{k=1}^{N_t}$, where $\ve{y}_k \in \mathbb{R}^{N_y}$ are noisy observations at time $t_k$. 

    \item \textbf{Prediction:} To use the estimated discrepancy and system state information as a correction mechanism for predictive use, embedding it into the nominal model so as to obtain improved forecasts of the system response under previously unseen excitations $\ve{f}^\ast(t)$. This step extends the estimation capability to predictive rectification, as required in digital twin applications.
\end{itemize}

\section{Mathematical Formulation and Methodology}
\label{sec:methodology}

\subsection{High-dimensional nominal computational model} \label{sec:met_1}

As outlined in \cref{sec:problemstatement}, the true structural dynamics include an additional restoring force term $\ve{g}\brc{\bar{\ve{u}}(t), \dot{\bar{\ve{u}}}(t)}$ that accounts for nonlinearities, higher-order theory effects, damping mechanisms, missing components, or damage.  While this term is essential for a fully faithful representation of the system, it is typically unavailable to the modeller or deliberately neglected to achieve a simplified computational model.  

The resulting \emph{nominal} computational model is therefore expressed as
\begin{equation}
    \matr{M} \ddot{\tilde{\ve{u}}}(t) + \matr{C} \dot{\tilde{\ve{u}}}(t) + \matr{K} \tilde{\ve{u}}(t) = \ve{f}(t) ,
    \label{eq:nomFE}
\end{equation}
where $\tilde{\ve{u}}(t) \in \mathbb{R}^{N}$ is the displacement vector of $N$ physical DOFs, $\matr{M}, \matr{C}, \matr{K} \in \mathbb{R}^{N\times N}$ are the known mass, damping, and stiffness matrices, and $\ve{f}(t) \in \mathbb{R}^{N}$ is the external force vector. Note that the discrepancy between the true response $\bar{\ve{u}}(t)$ in \eqref{eq:trueFE} and the nominal response $\tilde{\ve{u}}(t)$ in \eqref{eq:nomFE} arises solely from the omission of $\ve{g}$. This systematic difference constitutes the MFE. The central aim of this work is to estimate the effect of $\ve{g}$---interpreted as a latent force (LF) acting on the nominal system---from available measurements, and to embed this correction back into the nominal simulator to recover predictive fidelity. \cref{eq:nomFE} therefore serves as the starting point for our methodology.

\subsection{Modal reduction for computational feasibility} \label{sec:met_2}
A direct attempt to estimate the discrepancy term $\ve{g}$ in the full FE space is computationally prohibitive, since realistic discretizations lead to thousands of DOFs and would require handling a prohibitively large number of latent forces, increasing the burden of inference.
This motivates the use of a reduced-order representation---a modal reduction in this case---which provides a tractable space in which the latent MFEs can be inferred probabilistically.

We employ a \emph{modal reduction}, which projects the high-dimensional system onto a low-dimensional subspace spanned by a limited number of dominant vibration modes. 
This is accomplished by solving the generalized eigenproblem
\begin{equation}
    \matr{K} \ve{\Phi} = \matr{M} \ve{\Phi} \matr{\Omega}^2 ,
\end{equation}
where $\ve{\Phi} \in \mathbb{R}^{N\times m}$ contains the $m$ retained mode shapes and $\matr{\Omega} = \mathrm{diag}\brc{\omega_1, \ldots, \omega_m}$ is the diagonal matrix of natural frequencies.  
Approximating the displacement as $\tilde{\ve{u}}(t) \approx \ve{\Phi}\,\tilde{\ve{q}}(t)$ yields the reduced-order (uncoupled) modal equations:
\begin{equation}
    \ddot{\tilde{\ve{q}}}(t) + \matr{\Xi} \dot{\tilde{\ve{q}}}(t) + \matr{\Omega}^2 \tilde{\ve{q}}(t) = \ve{p}(t) ,
    \label{eq:modal}
\end{equation}
where $\tilde{\ve{q}}(t) \in \mathbb{R}^{m}$ are the modal coordinates, $\matr{\Xi}$ is the modal damping matrix (diagonal under proportional damping), and $\ve{p}(t) = \ve{\Phi}^\mathrm{T}\ve{f}(t)$ is the reduced forcing vector.  

The choice of $m \ll N$ ensures that the essential dynamics of the structure are preserved while reducing the dimension of the discrepancy term to a tractable scale. 
This reduction makes it possible to impose non-parametric GP priors on the latent forces in the modal domain, where the inference scales linearly in $m$. 
Although modal reduction is adopted here, the framework can accommodate other model-order reduction techniques \cite{ohayon2014variational,yang2017advanced,nguyen2008efficient,mignolet2013review}.

\subsection{Non-parametric probabilistic representation of model-form errors}\label{sec:met_3}

Once the nominal linear computational model has been projected into the reduced modal domain, the effects of the omitted restoring force $\ve{g}$ can be introduced in a computationally tractable way. 
Specifically, we model the MFE as an explicit time-dependent latent force (LF) $\ve{\eta}(t) \in \mathbb{R}^m$ acting on the reduced system, such that
\begin{equation}
    \ddot{\ve{q}}(t) + \matr{\Xi} \dot{\ve{q}}(t) + \matr{\Omega}^2 \ve{q}(t) + \ve{\eta}(t)= \ve{p}(t)  ,
    \label{eq:modal_MFE}
\end{equation}
where $\ve{q}(t)$ denotes the modal coordinates of the LF-augmented system and $\ve{p}(t)$ is the reduced external forcing.

The inclusion of $\ve{\eta}(t)$ causes the response of the LF-augmented system, $\ve{q}(t)$ (and hence $\ve{u}(t) \approx \matr{\Phi} \ve{q}(t)$), to deviate from the purely nominal modal response $\tilde{\ve{q}}(t)$ (and $\tilde{\ve{u}}(t)$ in the physical domain). If the LF model exactly captures the MFE, then the LF-augmented system response $\ve{u}(t)$ will mimic the true system response $\bar{\ve{u}}(t)$ perfectly in the physical domain. The difference between $\tilde{\ve{u}}(t)$ and $\bar{\ve{u}}(t)$ thus directly reflects the model-form error arising from omission of $\ve{g}$ in the nominal equations.

In contrast to parametric error models, which assume a predefined functional form, we adopt a \emph{non-parametric probabilistic representation}. 
Each component of $\ve{\eta}(t)$ is modeled as a zero-mean GP prior,
\begin{equation}
    \eta_i(t) \sim \mathcal{GP}\brc{0, \, \kappa\brc{t-t'; \ve{\theta}^{(i)}}} , \qquad i=1,\ldots,m ,
\end{equation}
with stationary covariance kernel $\kappa\brc{t-t'; \ve{\theta}^{(i)}}$ parameterized by hyperparameters $\ve{\theta}^{(i)}$. 
The full set of kernel hyperparameters is denoted $\ve{\theta} = \cbk{\ve{\theta}^{(1)}, \ldots, \ve{\theta}^{(m)}}$.

This non-parametric formulation has several advantages: 
\begin{itemize}
    \item It avoids restrictive parametric assumptions on the discrepancy structure, allowing flexible data-driven learning of MFE.  
    \item By working in the reduced modal space, the independence assumption between GPs becomes less restrictive, since modal equations are uncoupled. At the same time, the dimensionality of the latent force is small ($m \ll N$), making GP inference scalable.  
    \item Stationary covariance kernels admit equivalent state-space representations \cite{hartikainen2010kalman}, enabling convenient integration of the GP prior with the modal dynamics and efficient inference with linear-time complexity in time, in contrast to the cubic scaling of classical GP inference.  
\end{itemize}

Thus, by formulating the MFE as a GP latent force in the reduced modal domain, we obtain a probabilistic yet tractable representation of the unknown discrepancy, which forms the basis for the augmented state-space formulation described next.

\subsection{Augmented state-space formulation} \label{sec:met_4}

A key advantage of using stationary covariance kernels for the GP priors is that each independent GP can be equivalently represented by a linear stochastic differential equation (SDE). 
This allows the modally reduced structural dynamics \cref{eq:modal_MFE} and the GP latent force model to be combined into a single augmented state-space system.  

We define the augmented state vector
\begin{equation}
    \ve{z}(t) = \begin{bmatrix} \ve{q}(t) \\ \dot{\ve{q}}(t) \\ \ve{\eta}(t) \end{bmatrix} ,
\end{equation}
which includes the modal displacement $\ve{q}(t)$, velocity $\dot{\ve{q}}(t)$, and GP latent states $\ve{\eta}(t)$. 
The continuous-time state-space form of the augmented linear SDE can then be written as
\begin{align}
    \dot{\ve{z}}(t) &= \matr{A}_c(\ve{\theta}) \ve{z}(t) + \matr{B}_c \ve{p}(t) + \ve{w}(t; \ve{\theta}) , \label{eq:continuousSSM}
\end{align}
where $\ve{w}(t; \ve{\theta})$ is a zero-mean vector-valued Wiener process with covariance $\matr{Q}(\ve{\theta})$. The system matrices have the block form
\begin{equation}
    \matr{A}_c(\ve{\theta}) =
    \begin{bmatrix}
        \matr{0}_{m\times m}     & \matr{I}_{m}           & \matr{0}_{m\times m} \\
        -\matr{\Omega}^2         & -\matr{\Xi}            & -\matr{I}_{m}         \\
        \matr{0}_{m\times m}     & \matr{0}_{m\times m}   & \matr{F}(\ve{\theta})
    \end{bmatrix},
    \qquad
    \matr{B}_c =
    \begin{bmatrix}
        \matr{0}_{m\times m} \\
        \matr{I}_{m} \\
        \matr{0}_{m\times m}
    \end{bmatrix},
    \label{eq:block_Ac_Bc}
\end{equation}
where $\matr{\Omega}^2=\mathrm{diag}\brc{\omega_1^2,\ldots,\omega_m^2}$, $\matr{\Xi}=\mathrm{diag}\brc{\xi_1,\ldots,\xi_m}$ (assuming proportional damping), and $\matr{F}(\ve{\theta})$ encodes the GP state dynamics. Note $\matr{I}_n$ is an identity matrix of size $n \times n$, and $\zeros$ is a matrix of zeros.

For a Mat\'ern-$\tfrac{1}{2}$ (exponential) kernel,
\begin{align}
    \kappa \brc{t-t'; \ve{\theta}^{(i)}} = \alpha_i^2 \exp \brc{-\tfrac{|t-t'|}{\ell_i}} ,
\end{align}
the latent GP dynamics reduce to Ornstein–Uhlenbeck processes \cite{HartikainenPaper} with
\begin{align*}
    \matr{F}(\ve{\theta}) &= \mathrm{diag}\cbk{-1/\ell_1, -1/\ell_2, \ldots, -1/\ell_m} , \\
    \matr{Q}(\ve{\theta}) &= 
    \begin{bmatrix}
        \epsilon \,\matr{I}_{2m} & \matr{0}_{2m\times m} \\
        \matr{0}_{m\times 2m} & \mathrm{diag}\cbk{ 2\alpha_1^2/\ell_1,\, 2\alpha_2^2/\ell_2,\, \ldots,\, 2\alpha_m^2/\ell_m }
    \end{bmatrix},
\end{align*}
where $\epsilon>0$ is a small process-noise level for the $\brc{\ve{q},\dot{\ve{q}}}$ blocks (e.g., $\epsilon=10^{-12}$).

Discretizing \cref{eq:continuousSSM} with a sampling interval $\Delta t$ (using zero-order hold) yields the discrete-time process model
\begin{equation}
    \ve{z}_{k+1} = \matr{A}(\ve{\theta}) \ve{z}_k + \matr{B}\,\ve{p}_k + \ve{w}_k(\ve{\theta}) ,
    \label{eq:discreteSSM}
\end{equation}
where $\matr{A} = \mathrm{expm}\brc{\matr{A}_c(\ve{\theta}) \Delta t}$ is the state transition matrix and and $\matr{B} = \brc{\matr{A}-\matr{I}} \matr{A}_c^{-1}\matr{B}_c$. Note $\mathrm{expm}$ denotes matrix exponential.

The discrete-time process noise $\ve{w}_k(\ve{\theta})$ follows a multivariate Gaussian distribution with zero mean and covariance $\matr{Q}_d(\ve{\theta})$ given by
\begin{align}
    \matr{Q}_d(\ve{\theta}) &= \int_{0}^{\Delta t} 
        \matr{\Psi}_{a}(\Delta t-\tau)\,
        \matr{Q}(\ve{\theta})\,
        \matr{\Psi}_{a}^\top(\Delta t-\tau)\,
    d\tau , \label{eq:Qd_integral}
\end{align}
with $\matr{\Psi}_{a}(\tau) = \mathrm{expm}\brc{\matr{A}_c(\ve{\theta})\,\tau}$.
This integral can be evaluated efficiently using \emph{matrix-fraction decomposition}; see, e.g., Section 6.3 of \cite{sarkka2019applied} for derivations and implementation details.

This augmented state-space formulation provides a unified representation in which both the modal dynamics and the latent GP forces evolve jointly. 
It enables efficient inference of the MFE and system states using Kalman filtering, as described in the next subsection.

\subsection{Kalman filtering for joint state--discrepancy estimation} \label{sec:met_5}

With the discrete-time augmented state-space model in \cref{eq:discreteSSM}--\eqref{eq:Qd_integral}, the problem of estimating both the structural states and the latent MFE can be posed as a standard state estimation problem. 
The augmented state $\ve{z}_k \in \mathbb{R}^{3m}$ contains the modal displacement, velocity, and GP latent force components, and evolves linearly with Gaussian process noise.  

The available data consist of noisy measurements $\ve{y}_k \in \mathbb{R}^{N_y}$, in the form of displacements at a subset of DOFs $\ve{q}(t)$. 
These measurements are related to the augmented state through a linear measurement equation:
\begin{equation}
    \ve{y}_k = \matr{H} \matr{\Phi}\,\ve{z}_k + \ve{v}_k ,
    \label{eq:measurement}
\end{equation}
where $\matr{H}$ is an observation matrix, which extracts the rows of $\matr{\Phi}$ corresponding to the measured DOFs.
The measurement noise $\ve{v}_k$ is modeled as a zero-mean Gaussian vector, $\ve{v}_k \sim \mathcal{N}\brc{\ve{0},\,\matr{R}}$, with covariance $\matr{R}$ determined by sensor accuracy.

Given the linear Gaussian state-space model \crefrange{eq:discreteSSM}{eq:measurement}, joint inference of the modal states $\ve{q}_k$, velocities $\dot{\ve{q}}_k$, and latent GP forces $\ve{\eta}_k$ can be performed using a Kalman filter and Rauch--Tung--Striebel (RTS) smoother. 
This yields (a) separation of measurement noise $\ve{v}_k$ from the structural MFE $\ve{\eta}_k$, (b) uncertainty quantification for both structural states and latent discrepancy forces, and (b) optimal recursive estimation with linear computational complexity in the number of time steps.  

The GP kernel hyperparameters $\ve{\theta}$ are estimated by maximizing the posterior probability (MAP estimate):
\begin{equation}
\hat{\ve{\theta}} = \arg \max_{\ve{\theta}} \; p(\ve{\theta} \mid \ve{y}_{1:N_t}) \label{eq:maxopt}   
\end{equation}

In practice, this is achieved by running the Kalman filter for a candidate $\ve{\theta}$, with further details provided in Appendix \ref{appendix:MAP optimization of GP hyperparameters}. 

The Kalman filter and RTS smoother thus provide a principled mechanism for achieving the first objective of the problem statement: estimating the displacement trajectory of the true system and inferring the latent MFE term $\ve{\eta}(t)$ in a joint probabilistic framework. 
These inferred states and discrepancy estimates are subsequently used to construct a reusable correction model, as described next.

\subsection{Mesh-invariant mapping from states to discrepancies} \label{sec:met_6}
\label{sec:mesh_mapping}

The GPLFM framework provides posterior estimates of the latent MFE term $\ve{\eta}(t)$ at each time step, conditioned on measurements. 
While these estimates enable correction during the training phase, they cannot be used directly for predictive simulations under new excitations. 
To construct a reusable correction mechanism, we learn a functional mapping from mean values of modal states to the mean values of the MFEs.  

Specifically, we train a neural network model $\mathcal{N}_\psi$ to approximate the relation
\begin{equation}
    \ve{\eta}(t) \;\approx\; \mathcal{N}_{\ve{\psi}}\brc{\ve{q}(t),\, \dot{\ve{q}}(t)} ,
\end{equation}
using the smoothed mean estimates of $\cbk{\brc{\ve{q}_k,\dot{\ve{q}}_k},\ve{\eta}_k}_{k=1}^{N_t}$ obtained from the GPLFM inference stage as training data.

A crucial feature of this design is that the network operates entirely in the reduced modal space. 
Since modal coordinates are intrinsic to the structure and independent of the discretization density of the FE mesh, this representation ensures \emph{mesh invariance}: the trained surrogate $\mathcal{N}_\psi$ can be transferred to simulations of the same structure under different mesh resolutions without retraining.  
This property is particularly valuable in digital twin settings, where FE models may be refined or coarsened adaptively during the operational life of the structure.

\begin{figure}[t]
    \centering
    \includegraphics[width=0.85\linewidth]{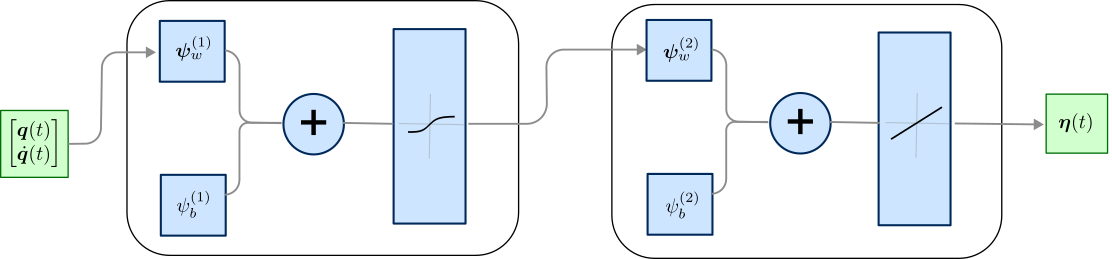}
    \caption{A two-layer neural network maps mean modal states $[\ve{q}(t),\,\dot{\ve{q}}(t)]$ to the mean latent discrepancy force $\ve{\eta}(t)$ in the reduced space. Hidden units use sigmoid activations, and the output is linear. The neural network is parameterized by $\ve{\psi} = [\ve{\psi}^{(1)^\top}_{w},\psi^{(1)^\top}_{b},\ve{\psi}^{(2)^\top}_{w},\psi^{(2)^\top}_{b}]$.}
    \label{fig:mapping_nn}
\end{figure}

We instantiate $\mathcal{N}_\psi$ as a compact two-layer neural network (with a single hidden layer) that maps modal states to modal latent forces (see \cref{fig:mapping_nn}). The input is the concatenated modal state vector $[\ve{q}^\top(t), \ \dot{\ve{q}}^\top(t)]^\top \in \mathbb{R}^{2m}$ and the output is $\ve{\eta}(t)\in\mathbb{R}^{m}$. Hidden units use sigmoid activations and the output layer is linear. Training data comprise the smoothed means $\{(\hat{\ve{q}}_k,\hat{\dot{\ve{q}}}_k),\hat{\ve{\eta}}_k\}_{k=1}^{N_t}$ obtained from the GPLFM stage. The network is trained by minimizing the mean-squared error using the ADAM optimizer~\cite{adam2014method}, with standard $z$–score normalization applied to both inputs and outputs, early stopping based on a held-out validation set, and a mild $\ell_2$ regularization term to mitigate overfitting. The same architecture and hyperparameters are consistently used across all examples presented in \cref{sec:numerical_examples}.

This mesh-invariant mapping provides a portable, data-driven correction model that can be easily embedded into the nominal reduced-order equations to obtain the rectified simulator described in the next subsection.

\subsection{Embedding the correction into the nominal computational model} \label{sec:met_7}

The mesh-invariant surrogate $\mathcal{N}_{\ve{\psi}}$, trained in the reduced modal space, provides a reusable correction model that predicts the latent MFE force from the instantaneous modal state. 
This surrogate can be embedded directly into the reduced-order nominal equations, yielding a \emph{rectified} nonlinear model:
\begin{equation} \label{eq:rectifiedeq}
     \ddot{\ve{q}}(t) + \matr{\Xi} \dot{\ve{q}}(t) + \matr{\Omega}^2 \ve{q}(t) + \mathcal{N}_{\ve{\psi}} \brc{\ve{q}(t), \dot{\ve{q}}(t)}
     = \ve{p}(t) 
\end{equation}

\cref{eq:rectifiedeq} preserves the physics-based structure of the nominal reduced-order system, while incorporating a black-box correction term inferred from data. 
This rectified model (see \cref{fig:mapping_nn}) operates as a standalone simulator; that is, once it is trained, $\mathcal{N}_\psi$ can be queried to provide corrective restoring forces from $[\ve{q}(t),\dot{\ve{q}}(t)]$ during time integration.

Given a new excitation $\ve{f}^\ast(t)$, the corresponding modal forcing $\ve{p}^\ast(t)=\ve{\Phi}^\top \ve{f}^\ast(t)$ is supplied to \cref{eq:rectifiedeq}, which is then integrated forward in time (using a numerical time-marching scheme such as fourth-order Runge-Kutta scheme). 
The reconstructed physical displacement response is obtained as
\begin{equation}
    \ve{u}^\ast(t) \;\approx\; \ve{\Phi}\,\ve{q}^\ast(t).
\end{equation}
This enables prediction of the structural response under unseen operating conditions, thereby achieving the second objective of the problem statement.

\paragraph{Digital twin perspective}
From a digital twin standpoint, the rectified model serves as a hybrid twin: the backbone dynamics are governed by the linear physics-based FE reduction, while the discrepancy is supplied by a data-driven, mesh-invariant surrogate. 
This ensures that the simulator remains partially interpretable while having good prediction ability, maintaining fidelity across different discretizations and excitation scenarios without repeated retraining.

\begin{algorithm}[H]
\caption{GPLFM-based joint MFE and state estimation}
\label{alg:gplfm}
\begin{algorithmic}
\Require FE model matrices $\brc{\matr{M},\matr{C},\matr{K}}$, force history $\ve{f}(t)$, displacement measurements $\mathcal{D}=\cbk{(t_k,\ve{y}_k)}_{k=1}^{N_t}$, sampling interval $\Delta t$
\Ensure Joint estimates $\cbk{\hat{\ve{q}}_k,\hat{\dot{\ve{q}}}_k,\hat{\ve{\eta}}_k}_{k=1}^{N_t}$, learned kernel hyperparameters $\hat{\ve{\theta}}$, mesh-invariant surrogate $\mathcal{N}_\psi$, and rectified simulator

\Statex \textbf{(A) Modal reduction \& nominal model (\cref{sec:met_2})}
\State Solve $\matr{K}\ve{\Phi}=\matr{M}\ve{\Phi}\matr{\Omega}^2$; select $m\ll N$ dominant modes to form $\ve{\Phi}\in\mathbb{R}^{N\times m}$ and $\matr{\Omega}=\mathrm{diag}\brc{\omega_1,\ldots,\omega_m}$
\State Compute reduced forcing $\ve{p}(t)=\ve{\Phi}^\top\ve{f}(t)$ and damping matrix $\matr{\Xi} \in \mathbb{R}^{m\times m}$
\State Obtain reduced nominal dynamics: $\ddot{\ve{q}}(t)+\matr{\Xi}\dot{\ve{q}}(t)+\matr{\Omega}^2\ve{q}(t)=\ve{p}(t)$

\Statex \textbf{(B) GP latent-force prior in modal space (\cref{sec:met_3})}
\State Place independent GP priors per modal DOF: $\eta_i(t)\sim\mathcal{GP}\!\brc{0,\kappa(t-t';\ve{\theta}^{(i)})}$, $i=1{:}m$
\State Choose stationary kernel (e.g., Mat\'ern-$\tfrac{1}{2}$) and collect hyperparameters $\ve{\theta}=\cbk{\ve{\theta}^{(1)},\ldots,\ve{\theta}^{(m)}}$

\Statex \textbf{(C) Augmented continuous-time SSM (\cref{sec:met_4})}
\State Form augmented state $\ve{z}=\sbk{\ve{q}^\top,\ \dot{\ve{q}}^\top,\ \ve{\eta}^\top}^\top\in\mathbb{R}^{3m}$
\State Build blocks:
\[
\matr{A}_c(\ve{\theta})=\begin{bmatrix}
\matr{0} & \matr{I} & \matr{0}\\
-\matr{\Omega}^2 & -\matr{\Xi} & -\matr{I}\\
\matr{0} & \matr{0} & \matr{F}(\ve{\theta})
\end{bmatrix},\quad
\matr{B}_c=\begin{bmatrix}\matr{0}\\ \matr{I}\\ \matr{0}\end{bmatrix},\quad
\matr{Q}(\ve{\theta})=\begin{bmatrix}
\epsilon\matr{I}_{2m} & \matr{0}\\
\matr{0} & \mathrm{diag}\brc{2\alpha_1^2/\ell_1,\ldots,2\alpha_m^2/\ell_m}
\end{bmatrix}
\]
\State Discretize the system with zero-order hold with $\Delta t$: $\matr{A}=\textrm{expm}\brc{\matr{A}_c\Delta t}$,\; $\matr{B}=(\matr{A}-\matr{I})\matr{A}_c^{-1}\matr{B}_c$
\State Compute $\matr{Q}_d(\ve{\theta})=\displaystyle\int_0^{\Delta t}\!\matr{\Psi}_a(\Delta t-\tau)\matr{Q}(\ve{\theta})\matr{\Psi}_a^\top(\Delta t-\tau)\,d\tau$, with $\matr{\Psi}_a(\tau)=\textrm{expm}\brc{\matr{A}_c\tau}$ (matrix-fraction decomposition; see \cite{sarkka2006thesis,sarkka2019applied})

\Statex \textbf{(D) Measurement model (\cref{eq:measurement})}
\State Set $\matr{H}$ based on the sensor locations $\Rightarrow$ pick rows of $\matr{\Phi}$ corresponding to the measured DOFs 
\State Set $\ve{y}_k=\matr{H}\matr{\Phi} \ve{z}_k+\ve{v}_k$, with $\ve{v}_k\sim\mathcal{N}\!\brc{\ve{0},\matr{R}}$

\Statex \textbf{(E) Hyperparameter learning (\cref{eq:maxopt} and \cref{appendix:MAP optimization of GP hyperparameters})}
\State Initialize $\ve{\theta}$ and prior $\mathcal{N}\!\brc{\ve{z}_0;\,\tilde{\ve{z}}_0,\matr{P}_{0|0}}$
\Repeat
  \State Run Kalman filter on $\cbk{\ve{y}_k}_{k=1}^{N_t}$ with current $\ve{\theta}$ to obtain predictive terms
  \State Accumulate $\log p\!\brc{\ve{y}_{1:N_t}\mid\ve{\theta}}=\sum_{k=1}^{N_t}\log p\!\brc{\ve{y}_k\mid\ve{y}_{1:k-1},\ve{\theta}}$
  \State Update $\ve{\theta}\leftarrow\ve{\theta}-\gamma\nabla_{\ve{\theta}}\!\big(-\log p(\ve{y}_{1:N_t}\mid\ve{\theta})-\log p(\ve{\theta})\big)$
\Until convergence to $\hat{\ve{\theta}}$

\Statex \textbf{(F) State and MFE inference}
\State With $\hat{\ve{\theta}}$, run Kalman filter \emph{and} RTS smoother to obtain $\cbk{\hat{\ve{z}}_{k|N_t},\matr{P}_{k|N_t}}_{k=1}^{N_t}$
\State Extract $\hat{\ve{q}}_k,\ \hat{\dot{\ve{q}}}_k,\ \hat{\ve{\eta}}_k$ from $\hat{\ve{z}}_{k|N_t}$
\end{algorithmic}
\end{algorithm}

\begin{algorithm}[H]
\caption{Neural network-based rectification}
\label{alg:rectified-simulation}
\begin{algorithmic}
\Require Mean of modal state $(\hat{\ve{q}}_k,\hat{\dot{\ve{q}}}_k)$ and MFE $\hat{\ve{\eta}}_k$ (obtained via \cref{alg:gplfm}), FE model matrices $\brc{\matr{M},\matr{C},\matr{K}}$, unseen force history $\ve{f}^\ast(t)$, sampling interval $\Delta t$
\Ensure Mesh-invariant surrogate $\mathcal{N}_\psi$, and rectified simulator

\Statex \textbf{(A) Modal reduction \& nominal model (\cref{sec:met_2})}
\State Solve $\matr{K}\ve{\Phi}=\matr{M}\ve{\Phi}\matr{\Omega}^2$; select $m\ll N$ dominant modes to form $\ve{\Phi}\in\mathbb{R}^{N\times m}$ and $\matr{\Omega}=\mathrm{diag}\brc{\omega_1,\ldots,\omega_m}$
\State Compute damping matrix $\matr{\Xi} \in \mathbb{R}^{m\times m} $ 
\Statex \textbf{(B) Mesh-invariant correction model}
\State Form training pairs using the means of $\big(\,(\hat{\ve{q}}_k,\hat{\dot{\ve{q}}}_k),\ \hat{\ve{\eta}}_k\,\big)$
\State Train neural network $\mathcal{N}_\psi:\brc{\hat{\ve{q}}_k,\hat{\dot{\ve{q}}}_k}\mapsto \hat{\ve{\eta}}_k$ 
f
\Statex \textbf{(C) Rectified simulation (second objective)}
\State Embed surrogate into reduced dynamics:
\[
\ddot{\ve{q}}(t)+\matr{\Xi}\dot{\ve{q}}(t)+\matr{\Omega}^2\ve{q}(t)+\mathcal{N}_\psi\!\brc{\ve{q}(t),\dot{\ve{q}}(t)}=\ve{p}^\ast(t)
\]
\State For a new load $\ve{f}^\ast(t)$, set $\ve{p}^\ast(t)=\ve{\Phi}^\top\ve{f}^\ast(t)$ and simulate the rectified system to obtain predictions $\cbk{\ve{u}^\ast(t_k)}$ with $\ve{u}^\ast\approx\ve{\Phi}\ve{q}^\ast$
\end{algorithmic}
\end{algorithm}

\begin{figure}[ht]
    \centering
    \includegraphics[width=0.95\linewidth]{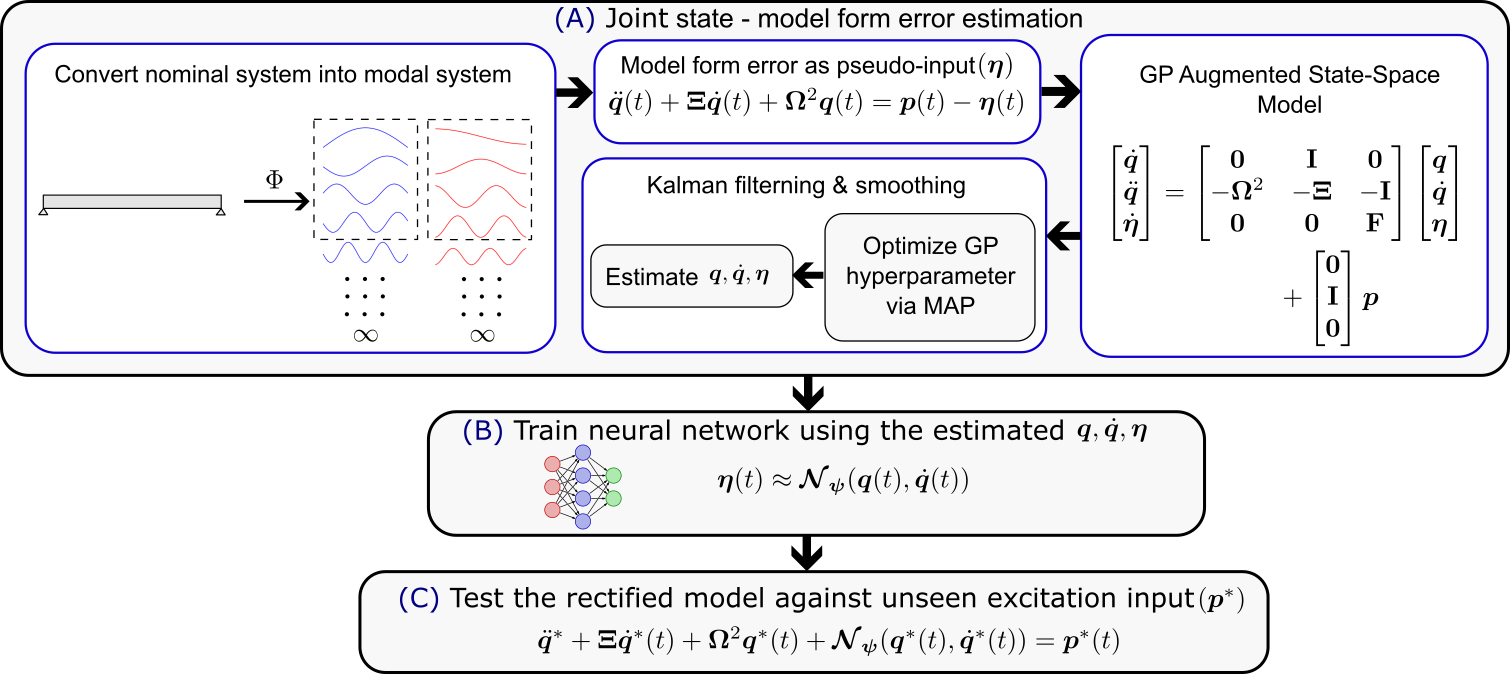}
    \caption{Rectification of the model discrepancy is done mainly in three stages as follows: (A) Estimate state $\ve{q}(t),\ve{\dot{q}}(t)$ and model form error $\ve{\eta}(t)$ (B) Train neural network using the estimated $\ve{q}(t),\ve{\dot{q}}(t),\ve{\eta}(t)$ (C) Test the rectified model against unseen external forcing function ($\ve{p}^*(t) = \Phi^\top \ve{f}^*(t)$)}
    \label{fig:workflow}
\end{figure}

\section{Illustrative Examples} \label{sec:numerical_examples}
To evaluate the proposed framework for MFE inference and prediction-rectification, five numerical case studies are considered, each representing a distinct source of modeling discrepancy: (i) mismatch in beam theory, (ii) damping misspecification, (iii) boundary condition misspecification, (iv) nonlinear constitutive law, and (v) local structural damage. Together, these examples demonstrate the framework's ability to estimate latent discrepancies, train mesh-invariant surrogates, and improve predictive accuracy under both noiseless and noisy measurements.

\subsection{Numerical Experiment Design} \label{sec:DOE}
The case studies are organized into two groups: the first four consider a prismatic simply supported beam discretized with 200 DOFs, while the fifth focuses on a 120-DOF ASCE benchmark building model. Across all cases, the framework is tested for its ability to estimate discrepancies and rectify predictions when exposed to both training excitations and unseen test inputs.

\subsubsection*{Beam Models (Examples 1--4)}
The homogeneous beam (see \cref{fig:beam_setup}) has length $L=10$~m and rectangular cross-section ($b=0.4$~m, $h=0.5$~m), with linear elastic material properties: Young's modulus $E=200$~GPa, Poisson's ratio $\nu=0.3$, and mass density $\rho=7800~\mathrm{kg/m^3}$. Nine displacement sensors are equally spaced from 1~m to 9~m along the span. Dynamic point loads are applied at 2~m, 5~m, and 8~m. During training, the loads are sinusoidal with amplitudes of 105~kN at frequencies of 37, 25, and 50~rad/s, while during testing, amplitudes are reduced to 90~kN and frequencies shifted to 30, 20, and 60~rad/s. In Example~4 (hyperelastic material), the load amplitudes are increased by an order of magnitude (1050~kN training, 900~kN testing) to elicit nonlinear responses.

\begin{figure}[H]
    \centering

\begin{tikzpicture}[box/.style={rectangle, draw, rounded corners, fill=blue!20, text centered, minimum height=1cm, minimum width=2.5cm,text width=4cm},
arrow/.style={-{Latex}, thick}
]

\draw[draw=black, fill=gray!10] (10,5) rectangle ++(14,0.7);
\draw[draw=black, fill=gray!50] (9.8,4.8) -- (10,5) -- (10.2,4.8) -- cycle;
\draw[draw=black, fill=gray!50] (10+13.8,4.8) -- (10+14,5) -- (10+14.2,4.8) -- cycle;

\draw[arrow] (10+2*1.4,7) -- (10+2*1.4,5.7);
\draw[arrow] (10+5*1.4,7) -- (10+5*1.4,5.7);
\draw[arrow] (10+8*1.4,7) -- (10+8*1.4,5.7);
\draw [->,blue] (10+12.9,6.2) to [out=160,in=100] (10+9*1.4,5.81);
\draw[<->] (10,4.5) -- (10+14,4.5);

\node[] at (10+2*1.4,7.4) {$105\sin{(37t)}$ \si{kN}};
\node[] at (10+5*1.4,7.4) {$105\sin{(25t)}$ \si{kN}};
\node[] at (10+8*1.4,7.4) {$105\sin{(50t)}$ \si{kN}};
\node[text=blue] at (10+13.5,6.2) {sensor};
\node[] at (10+7,4.7) {10 \si{m}};

\node[] at (10+1.5*1.4,3.3) {$E=200\times10^9$ \si{N/m^2},};
\node[] at (10+3.8*1.4,3.3) {$\rho=7800 \ \si{kg/m^3}$,};
\node[] at (10+5.3*1.4,3.3) {$\nu=0.3$,};
\node[] at (10+6.5*1.4,3.3) {$b=0.4$ \si{m},};
\node[] at (10+7.8*1.4,3.3) {$h=0.5$ \si{m},};
\node[] at (10+9*1.4,3.3) {$L=10$ \si{m}};

\filldraw[fill=red] (10+1.4,5.7) circle (2pt);
\filldraw[fill=red] (10+2*1.4,5.7) circle (2pt);
\filldraw[fill=red] (10+3*1.4,5.7) circle (2pt);
\filldraw[fill=red] (10+4*1.4,5.7) circle (2pt);
\filldraw[fill=red] (10+5*1.4,5.7) circle (2pt);
\filldraw[fill=red] (10+6*1.4,5.7) circle (2pt);
\filldraw[fill=red] (10+7*1.4,5.7) circle (2pt);
\filldraw[fill=red] (10+8*1.4,5.7) circle (2pt);
\filldraw[fill=red] (10+9*1.4,5.7) circle (2pt);
\end{tikzpicture}

\caption{Experimental setup: simply supported beam with point input force given at three locations 2 \si{m}, 5 \si{m} and 8 \si{m} from left side of the beam. Sensors are placed at equidistance starting from location 1 \si{m} up to 9 \si{m} of the beam and are represented by a red dot}
\label{fig:beam_setup}
\end{figure}

\begin{figure}
    \centering
    \includegraphics[width=0.75\linewidth]{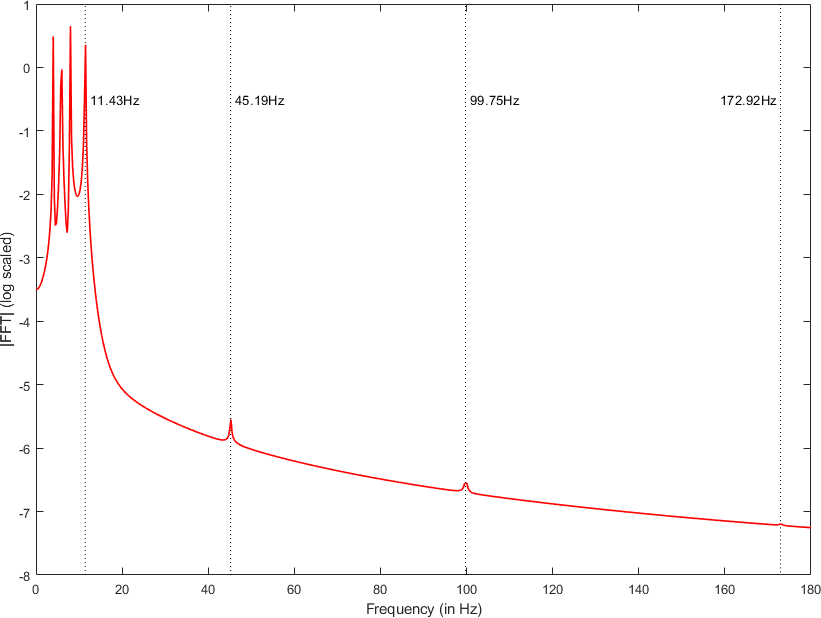}
    \caption{Averaged log–magnitude FFT of displacement (and rotation) from the nominal linear beam at the nine sensor locations. Peaks at the first four structural natural frequencies are shown in dotted vertical lines; mode~4 is weak but discernible, while no clear peaks are observed beyond mode~4.}
    \label{fig:mean_of_fft}
\end{figure}

The nominal beam model is projected onto the first four vibration modes to construct the reduced-order basis. This choice is motivated by the frequency content of the nominal linear system response. \cref{fig:mean_of_fft} shows the averaged log-magnitude FFT of displacement and rotation signals recorded at the sensor locations. The plot reveals that the first mode is strongly excited by the applied loading, while the second and third modes are only weakly excited but remain clearly discernible. The fourth mode contributes very little energy, yet its presence is non-negligible. Beyond the fourth mode, no distinct peaks corresponding to structural natural frequencies are observed, indicating negligible dynamic contribution. On this basis, retaining the first four modes is sufficient to capture the essential dynamics of the system without introducing unnecessary computational burden.

\subsubsection*{120-DOF Shear-Building Model (Example 5)}
The final case uses a 120-DOF shear building subjected to narrow-band Gaussian excitation. 
Training and testing inputs consist of different realizations of the narrow-band Gaussian excitation. 
As with the beams, we perform a modal reduction and retain the first $m$ lateral modes identified from the averaged log–magnitude FFT of the measured outputs; in our implementation $m=9$, which captures the visible peaks from the averaged response spectrum.
This example departs from the beam family and evaluates whether the framework can infer a \emph{localized} stiffness loss (damage) and transfer the mesh-invariant correction to predictive simulations under a different excitation input.

\subsubsection*{Measurement Type and Noise}
All examples use displacement measurements for inference and rectification. Displacements yield better conditioning and allow consistent comparison of the framework across examples.

Robustness is assessed by adding white Gaussian noise to the true displacement signals. Noise is sampled from i.i.d.\ Gaussian distribution $\mathcal{N}(0,\sigma_n^2)$ with standard deviation
\begin{equation}
\sigma_{n} = \frac{p}{100} \times \frac{1}{N_m}\sum_{i=1}^{N_m}\sqrt{\frac{1}{N_t}\sum_{k=1}^{N_t} \bar{u}_{i}^2(t_k)},
\end{equation}
where $p$ is the noise percentage, $N_m$ the number of sensors, and $N_t$ the number of time steps. Two noise scenarios are considered: 1\% and 5\%.

\subsubsection*{Neural Network Configuration}
For all examples, the mapping from mean modal states to the mean modal latent forces is realized using a fully connected feed-forward neural network with two hidden layers (see \cref{fig:mapping_nn}). The number of neurons in the hidden layer is selected through cross-validation to minimize the validation mean-squared error, typically ranging between 20 and 100 neurons. 

\subsubsection*{Validation Strategy}
The \emph{true} discrepancy is obtained by evaluating the residual of the nominal model when driven by the true response and projecting it into the modal subspace:
\begin{equation}
    \ve{\eta}(t) = \matr{\Phi}^\top \sbk{\ve{f}(t) - \matr{M}\ddot{\bar{\ve{u}}}(t) - \matr{C}\dot{\bar{\ve{u}}}(t) - \matr{K}\bar{\ve{u}}(t)}
\end{equation}
This ensures a consistent comparison between the estimated latent forces and the ground-truth discrepancy in the reduced modal domain.

The performance of the proposed methodology is validated along three lines: (i) accuracy of state estimation for displacements and rotations across all physical DOFs, (ii) fidelity of discrepancy recovery in modal space, and (iii) predictive accuracy of the rectified model under unseen excitations. Quantitative assessment is based on normalized mean squared error (NMSE) in percentage:

\begin{align}
e \brc{\ve{p}, \ve{\hat{p}}} = \sbk{\frac{1}{n} \frac{1}{N_t} \; \sum_{i=1}^{n}\frac{1}{{\sigma_{p_i}}^2} \; \sum_{k =1}^{N_t}(p_{i}(t_k) - \hat{p}_{i}(t_k))^2 }\times 100 \%
\end{align}
In the above definition, $n$ denotes the dimension of the vector $\ve{p}(t)$, $N_t$ represents the total number of signal samples, $p_i(t_k)$ and $\hat{p}_i(t_k)$ correspond to the $i$th component of the true and the estimated vector signals, respectively. Additionally, $\sigma_{p_i}^2$ signifies the variance of the component signal $p_i(t)$.

\subsection{Example 1: Beam theory mismatch}
The first example investigates MFEs that arise when a lower-order beam theory is used in place of a higher-fidelity description. The \emph{true} structural response is governed by Timoshenko beam theory with shear deformation and rotary inertia effects, while the \emph{nominal} model adopts the Euler--Bernoulli (E-B) beam theory. 

\paragraph{True model}  
The linear elastic Timoshenko beam with viscous damping is described by
\begin{subequations}
    \begin{align}
        \rho A \ddot{\bar{w}}(x,t) + c_w \dot{\bar{w}}(x,t) - kGA \brc{\tfrac{\pd \bar{w}(x,t)}{\pd x} - \beta(x,t)} &= p(x,t), \quad 0 \le x \le L,\; 0 \le t \le T, \\
        \rho I \ddot{\beta}(x,t) + c_\beta\dot{\beta}(x,t) - EI \tfrac{\pd^2 \beta(x,t)}{\pd x^2} - kGA \brc{\tfrac{\pd \bar{w}(x,t)}{\pd x} - \beta(x,t)} &= 0, \quad 0 \le x \le L,\; 0 \le t \le T 
    \end{align}
\end{subequations}
where $\bar{w}(x,t)$ is the transverse displacement, $\beta(x,t)$ is the cross-sectional rotation, $G = \frac{E}{2(1+\nu)}$ is the shear modulus, $A$ is the cross-sectional area, $I$ is the second moment of area, $k$ is the shear correction factor, $c_w$ and $c_\beta$ are the viscous damping coefficient per unit length. The system is subject to zero initial conditions (ICs) and simply supported boundary conditions (BCs):
\begin{align}
\begin{split}
    \bar{w}(x,0) = 0, &\quad \dot{\bar{w}}(x,0) = 0, \\
    \bar{w}(0,t) = 0,\; \bar{w}(L,t) = 0, &\quad \tfrac{\pd \beta}{\pd x}(0,t) = 0,\; \tfrac{\pd \beta}{\pd x}(L,t) = 0
\end{split}
\end{align}

\paragraph{Nominal model}  
The E-B beam model retains only the transverse displacement $\tilde{w}(x,t)$:
\begin{align}
    \rho A \ddot{\tilde{w}}(x,t) + c_w \dot{\tilde{w}}(x,t) + EI \tfrac{\pd^4 \tilde{w}}{\pd x^4} = p(x,t), \quad 0 \le x \le L,\; 0 \le t \le T,
\end{align}
with ICs and BCs
\begin{align}
\begin{split}
    \tilde{w}(x,0) = 0, &\quad \dot{\tilde{w}}(x,0) = 0, \\
    \tilde{w}(0,t) = 0,\; \tilde{w}(L,t) = 0, &\quad \tfrac{\pd^2 \tilde{w}}{\pd x^2}(0,t) = 0,\; \tfrac{\pd^2 \tilde{w}}{\pd x^2}(L,t) = 0.
\end{split}
\end{align}
By neglecting shear deformation and rotary inertia, the nominal model enforces that cross-sections remain orthogonal to the neutral axis. This simplification reduces computational cost but introduces systematic bias in both displacement and rotation whenever shear effects are appreciable.

\paragraph{Discrepancy modeling}  
Within the proposed GPLFM framework, the omitted shear contributions are represented as latent forces in the reduced modal domain (four modes; see \cref{sec:DOE}). This allows the nominal E-B beam dynamics to be reconciled with the Timoshenko beam response during training and enhances predictions under unseen test excitations.

\begin{figure}[H]
    \centering
    \includegraphics[width=1\linewidth]{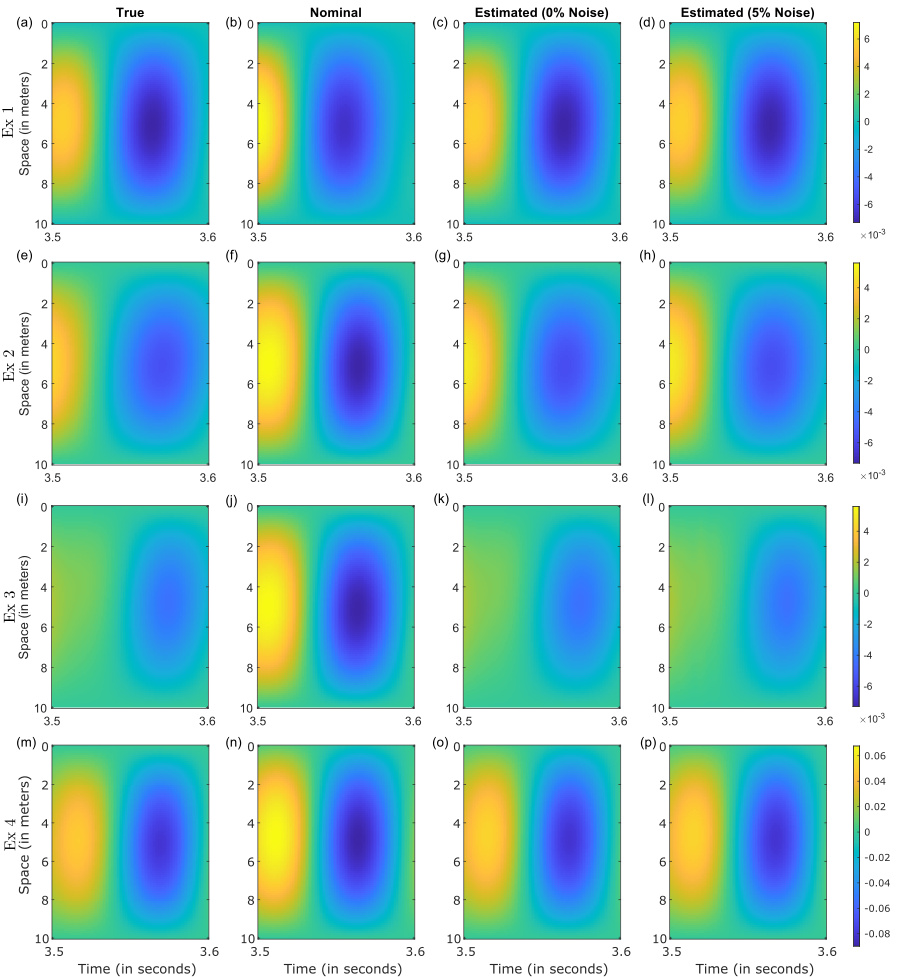}
    \caption{Spatio–temporal displacement \emph{estimation} on the full beam for a representative window (3.5–3.6~s). 
    Grid of panels: columns show (i) true displacement (Timoshenko), (ii) nominal displacement (model as specified in each example), (iii) GPLFM estimate (0\% noise), (iv) GPLFM estimate (5\% noise). 
    Rows correspond to Examples~1–4: (1) beam theory mismatch, (2) damping misspecification, (3) boundary condition misspecification, (4) nonlinear constitutive law. Color scale is consistent along rows.}
    \label{fig:all_examples_est}
\end{figure}

\paragraph{Results} 
\cref{fig:all_examples_est} (first row) compares the spatio-temporal displacement field across the beam span over a representative time window (3.5–3.6 s) for the true Timoshenko model response, the nominal E-B model prediction, and the GP-augmented-nominal-model estimates under 0\% and 5\% noise. In the noiseless case, the augmented model reproduces the true spatio–temporal field nearly exactly, while at 5\% noise, small fluctuations appear but the dominant displacement patterns remain well captured.

Quantitative errors for displacement and rotation inference appear in \cref{tab:NMSE_for_example_1_to_4} (first row). Across all noise levels, the GP-augmented nominal model reduces estimation NMSE by more than 99\% relative to the nominal E-B baseline.

\begin{figure}[H]
    \centering
    \includegraphics[width=0.75\linewidth]{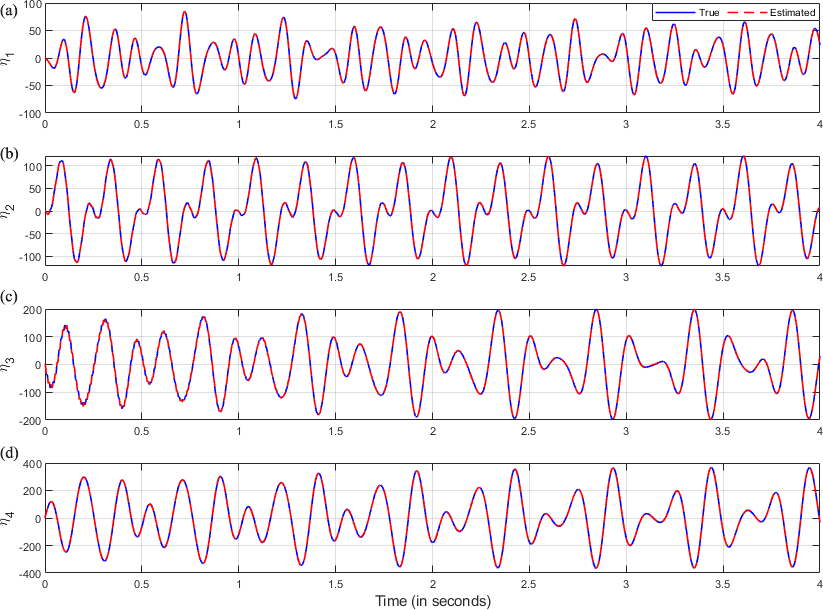}
    \caption{Example 1 (Beam theory misspecification): Comparison of inferred modal latent forces (red dashed) and ground-truth modal discrepancies (blue solid) for the four retained modes under 0\% noise.
    The GPLFM accurately recovers the latent forces corresponding to the omitted shear-deformation effects, reconciling the nominal Euler--Bernoulli model with the true Timoshenko beam response.}
    \label{fig:example1_discrepancy_est_noise0}
\end{figure}

\begin{figure}[H]
    \centering
    \includegraphics[width=0.85\linewidth]{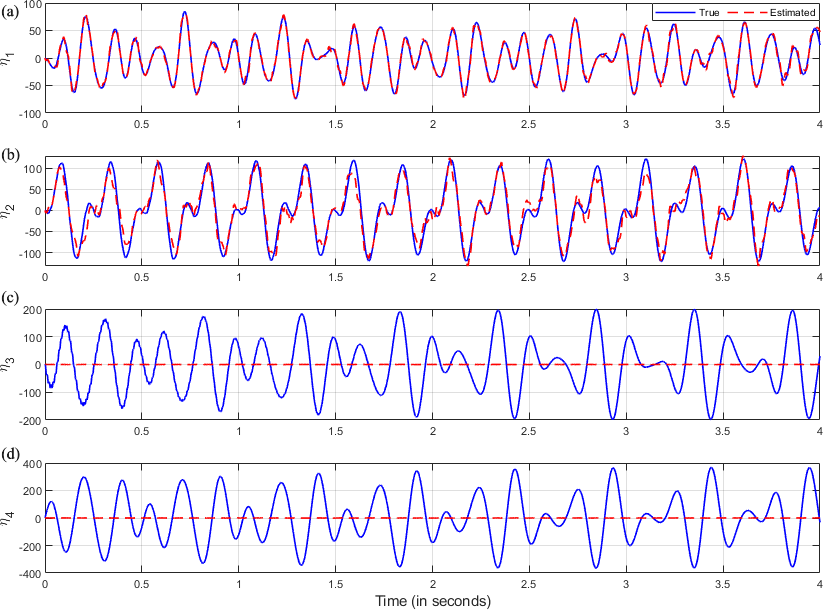}
    \caption{Example 1 (Beam theory misspecification): Inferred modal latent forces (red dashed) versus ground-truth discrepancies (blue solid) for 1\% measurement noise.
    The dominant first mode latent force remains accurately reconstructed, the second mode latent force has minor deviations, while those of higher modes (3 and 4) are near-zero due to their response contribution falling below the noise floor.}
    \label{fig:example1_discrepancy_est_noise1}
\end{figure}

\begin{figure}[H]
    \centering
    \includegraphics[width=0.85\linewidth]{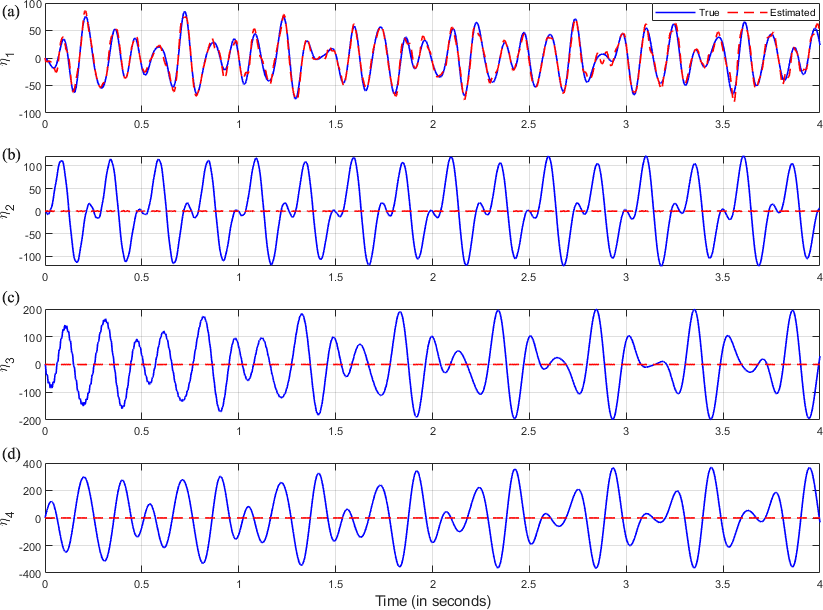}
    \caption{Example 1 (Beam theory misspecification): Inferred modal latent forces (red dashed) versus ground-truth discrepancies (blue solid) for 5\% measurement noise.
    The GPLFM continues to capture the first mode latent force associated with shear deformation, whereas higher-order modes (2–4) degrade to zero due to their low energy content and unidentifiability under noisy measurements.}
    \label{fig:example1_discrepancy_est_noise5}
\end{figure}

\cref{fig:example1_discrepancy_est_noise0,fig:example1_discrepancy_est_noise1,fig:example1_discrepancy_est_noise5} compare the inferred modal latent forces with the ground-truth shear-induced discrepancies (blue = true, red dashed = inferred). At 0\% noise, all four modes are faithfully reconstructed. With increasing noise, higher-mode discrepancies (particularly modes 2, 3, and 4) diminish relative to the noise floor, while the dominant first-mode discrepancy remains robustly captured.

\begin{figure}[H]
    \centering
    \includegraphics[width=1\linewidth]{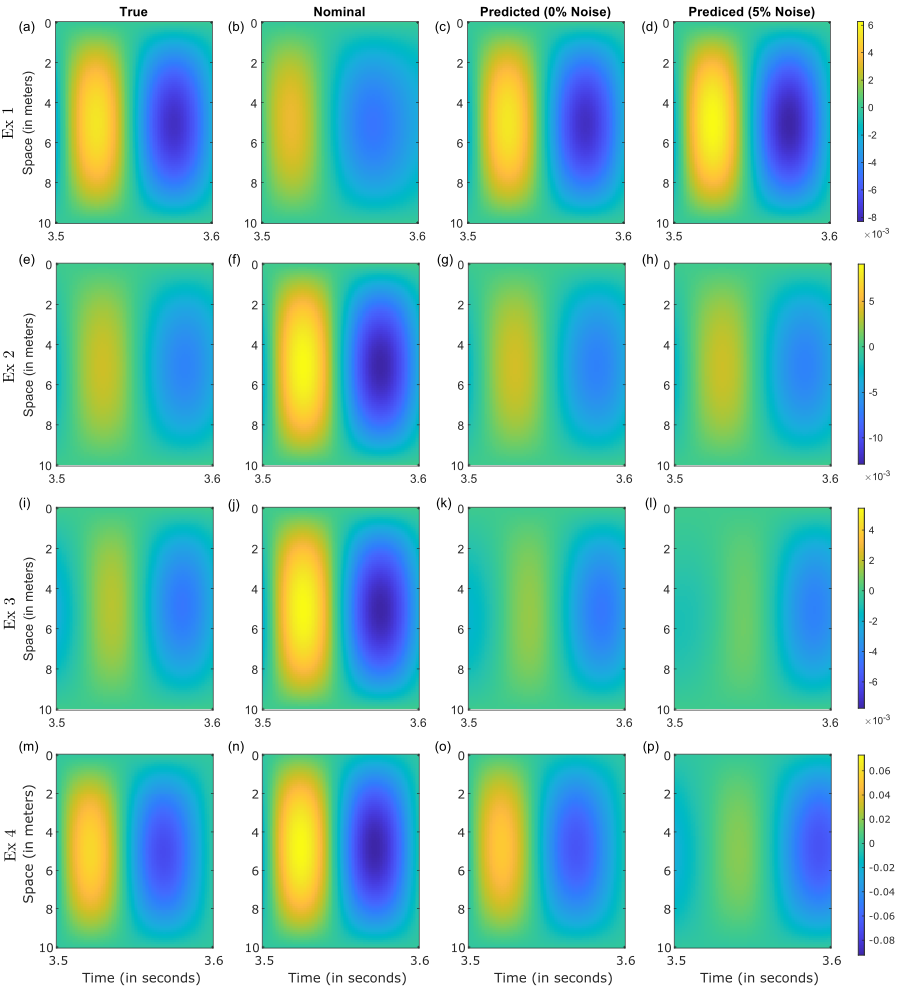}
    \caption{Spatio–temporal displacement \emph{prediction} under unseen sinusoidal inputs.
    Columns show (i) true displacement, (ii) nominal displacement, (iii) rectified prediction with 0\% noise training, and (iv) rectified prediction with 5\% noise training. 
    Rows correspond to Examples~1–4 as in \cref{fig:all_examples_est}.}
    \label{fig:all_examples_pred}
\end{figure}

\cref{fig:all_examples_pred} (first row) shows prediction under unseen test loading (specifications in \cref{sec:DOE}). The rectified model consistently outperforms the nominal E-B model. As measurement noise increases, accuracy degrades---as expected---because latent-force inference becomes less precise; nevertheless \cref{tab:NMSE_for_example_1_to_4} shows large NMSE reductions over the nominal NMSE.

\begin{table}[H]
\centering
\caption{Normalized mean squared error (NMSE, \%) for inference (estimation) and rectified prediction across Examples 1--4. 
Baseline nominal NMSEs are shown in columns 2 and 6. Values in parentheses indicate percentage reduction relative to the baseline.}
\label{tab:NMSE_for_example_1_to_4}
\begin{tabular}{|c|c|c|c|c|c|c|c|c|}
\hline
 & & \multicolumn{3}{c|}{\textbf{Inference NMSE $e(\bar{\ve{u}}, \ve{u})$}} & & \multicolumn{3}{c|}{\textbf{Rectified prediction $e(\bar{\ve{u}}^\ast, \ve{u}^\ast)$ }} \\
\cline{3-5}\cline{7-9}
\textbf{Ex} & $e(\bar{\ve u}, \tilde{\ve u})$ & Noiseless & 1\% & 5\% & $e(\bar{\ve{u}}^\ast, \tilde{\ve{u}}^\ast)$ & Noiseless & 1\% & 5\% \\
\hline
1 & 9.6 & 0.0034 (100.0) & 0.0067 (99.9) & 0.021 (99.8) & 10 & 0.16 (98.4) & 0.32 (97.0) & 0.98 (90.7) \\
\hline
2 & 6.6 & 0 (100.0) & 0.0031 (99.9) & 0.02 (99.7) & 83 & 0.13 (99.8) & 0.23 (99.7) & 0.31 (99.6) \\
\hline
3 & 140 & 0.063 (99.9) & 0.071 (99.9) & 0.33 (99.8) & 240 & 6.6 (97.3) & 11 (95.4) & 24 (90.2) \\
\hline
4 & 99 & 3.9 (96.1) & 3.8 (96.1) & 3.8 (96.1) & 100 & 4.4 (95.7) & 9.4 (90.9) & 61 (41.3) \\
\hline
\end{tabular}
\end{table}


\subsection{Example 2: Damping misspecification}
This example investigates MFEs that arise due to an incorrect damping representation. The \emph{true} structural dynamics are governed by the Timoshenko beam theory with Rayleigh damping, while the \emph{nominal} model assumes the same structural form but employs modal damping. The mismatch between the Rayleigh-damped and modally damped formulations introduces systematic phase and amplitude errors that the proposed GPLFM aims to capture through latent forces.

\paragraph{True model}  
The Rayleigh damping model represents the viscous damping matrix as a linear combination of the mass and stiffness matrices:
\begin{equation}
    \bar{\matr{C}} = \alpha \bar{\matr{M}} + \beta \bar{\matr{K}}
\end{equation}
where $\alpha$ and $\beta$ are the mass- and stiffness-proportional damping coefficients. These coefficients are identified by matching the target damping ratios $\zeta_i$ and $\zeta_j$ of two selected modes (typically the first and second modes) using the relations:
\begin{align}
\begin{split}
    \zeta_i &= \dfrac{1}{2}\brc{\dfrac{\alpha}{\omega_i}+\beta \omega_i}  \\
    \zeta_j &= \dfrac{1}{2}\brc{\dfrac{\alpha}{\omega_j}+\beta \omega_j}
\end{split}
\end{align}
where $\omega_i$ denotes the natural frequency (rad/s) of the $i^{\text{th}}$ mode.  
By construction, the Rayleigh model reproduces the target damping ratios exactly for these two modes, but the damping ratios for the remaining modes are only approximate and frequency-dependent.

\paragraph{Nominal model.}  
The nominal system uses a modal damping representation in which each retained vibration mode is assigned 
its own independent damping ratio $\tilde{\zeta}_i$. The damping matrix in modal coordinates is therefore diagonal:
\begin{equation}
    \tilde{\matr{\Xi}} = \mathrm{diag}\cbk{2\tilde{\zeta}_1\omega_1, \;2\tilde{\zeta}_2\omega_2, \ldots,\; 2\tilde{\zeta}_m\omega_m },
\end{equation}
where $m=4$ corresponds to the number of retained vibration modes. Although this approach simplifies the system representation, it fails to capture frequency-coupled damping 
effects that are implicit in the Rayleigh formulation, resulting in MFEs during both transient and steady-state response.

\paragraph{Discrepancy modeling}
Within the proposed GPLFM framework, the MFEs arising from damping misspecification are treated as latent forces 
acting in the reduced modal domain. Each latent force $\eta_i(t)$ compensates for the mismatch between 
Rayleigh and modal damping for the $i^{\text{th}}$ mode. This flexible representation aligns the nominal modal dynamics 
with the true Rayleigh-damped system during training and enables improved predictive performance under unseen excitations.

\paragraph{Results} 
\cref{fig:all_examples_est} (second row) compares the spatio–temporal displacement fields for the true Rayleigh-damped system, 
the nominal modally damped model, and the GP-augmented nominal estimates over the representative window (3.5–3.6 \si{s}). 
Under noiseless conditions, the augmented model reproduces the true field with near-perfect accuracy. 
Even with 5\% measurement noise, the key spatial and temporal patterns remain well preserved.  

Quantitative performance metrics are summarized in \cref{tab:NMSE_for_example_1_to_4} (second row). 
Across all noise levels, the GP-augmented nominal model achieves over 99\% reduction in displacement and rotation NMSE 
relative to the nominal baseline, demonstrating robust state inference.

\begin{figure}[H]
    \centering
    \includegraphics[width=0.75\linewidth]{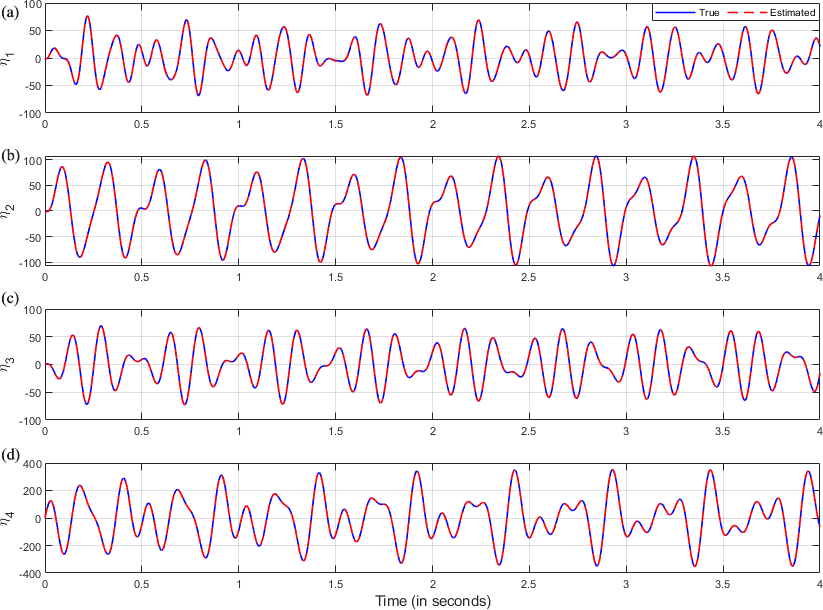}
    \caption{Example 2 (Damping misspecification): Comparison of inferred modal latent forces (red dashed) and ground-truth damping discrepancies (blue solid) for the four retained modes under 0\% noise. 
    The GPLFM successfully reproduces all modal latent forces, effectively compensating for the phase and amplitude errors introduced by the mismatch between Rayleigh and modal damping.}
    \label{fig:example2_discrepancy_est_noise0}
\end{figure}

\begin{figure}[H]
    \centering
    \includegraphics[width=0.75\linewidth]{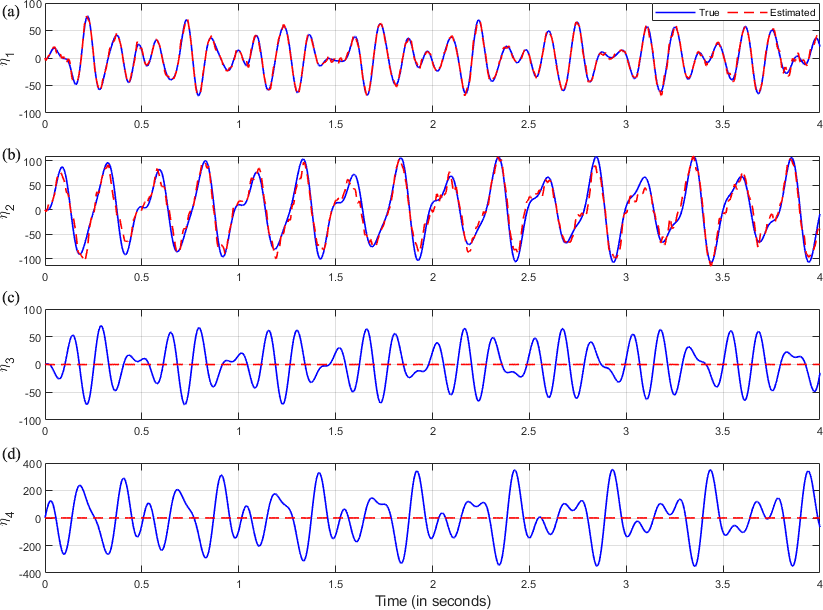}
    \caption{Example 2 (Damping misspecification): Inferred modal latent forces (red dashed) versus ground-truth damping discrepancies (blue solid) for 1\% measurement noise. 
    The dominant first modal latent force remains accurately reconstructed, whereas the second mode shows mild discrepancy, and higher modes (3 and 4) are unobservable below the noise floor.}
    \label{fig:example2_discrepancy_est_noise1}
\end{figure}

\begin{figure}[H]
    \centering
    \includegraphics[width=0.75\linewidth]{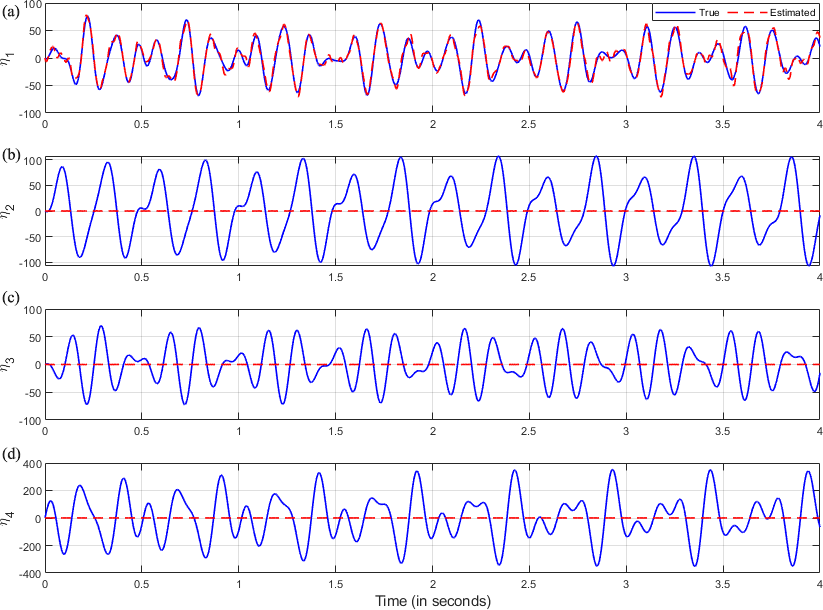}
    \caption{Example 2 (Damping misspecification): Inferred modal latent forces (red dashed) versus ground-truth damping discrepancies (blue solid) for 5\% measurement noise. 
    The first modal latent force continues to capture the essential damping mismatch, while the higher-order modes (2, 3, and 4) decay toward zero due to loss of identifiability below the noise threshold.}
    \label{fig:example2_discrepancy_est_noise5}
\end{figure}

The inferred modal latent forces are compared with the ground-truth damping discrepancies 
in \Cref{fig:example2_discrepancy_est_noise0,fig:example2_discrepancy_est_noise1,fig:example2_discrepancy_est_noise5}. 
At 0\% noise, all four modal latent forces closely match the true discrepancies. 
However, as the measurement noise increases, the estimated latent forces corresponding to the higher modes (modes 3 and 4) progressively deteriorate and often converge to nearly zero. 
This behavior arises because the higher modal responses contribute only marginally to the overall structural dynamics, as evidenced by their weak amplitudes in the output FFT (\cref{fig:mean_of_fft}). 
With increasing noise, these small-amplitude modal responses fall below the noise floor, causing their signal-to-noise ratio to drop drastically. 
Consequently, the GPLFM cannot retrieve meaningful latent forces for these modes, leading to poor or null estimation of higher-order latent components.

Finally, \cref{fig:all_examples_pred} (second row) illustrates rectified predictions under unseen sinusoidal excitations. 
Although prediction accuracy decreases slightly with higher noise due to degraded latent-force inference, the augmented model continues to outperform its nominal counterpart by a wide margin, 
as confirmed by the NMSE reductions in \cref{tab:NMSE_for_example_1_to_4}.


\subsection{Example 3: Misspecified boundary condition}
This example examines MFEs arising from an incorrect specification of boundary conditions, a common source of model discrepancy in structural dynamics. The \emph{true} system corresponds to a simply supported beam with a finite rotational restraint at its right end, while the \emph{nominal} system assumes an idealized simply supported beam with freely rotating ends. Such discrepancies occur frequently in realistic scenarios, as often physical supports are never perfectly free of rotational stiffness. The resulting MFE is spatially \emph{localized} near the boundary, yet---as shown below---the proposed GPLFM is able to infer its effect accurately even when working with global modal bases constructed from the nominal system, which do not explicitly capture this local stiffness.

\paragraph{True model}
The true beam is modeled using Timoshenko beam theory and includes a linear rotary spring of stiffness $k_\theta = 2\times10^8 \si{N\; m/rad}$ attached at the right end. The governing boundary conditions (BCs) are
\begin{equation}
   \bar{w}(0,t) = 0,\quad \bar{w}(L,t) = 0, \qquad \tfrac{\pd \beta}{\pd x}(0,t) = 0,\quad \tfrac{\pd \beta}{\pd x}(L,t) = \dfrac{-k_\theta \beta}{EI} 
\end{equation}
where $\bar{w}(x,t)$ and $\beta(x,t)$ denote the transverse displacement and rotation fields, respectively. This configuration introduces an additional rotational stiffness at $x=L$, thereby shifting the mode shapes and natural frequencies relative to the nominal simply supported case.

\paragraph{Nominal model}
The nominal beam, also formulated using Timoshenko beam theory, neglects the localized rotary stiffness and assumes classical simply supported conditions:
\begin{equation}
   \bar{w}(0,t) = 0,\quad  \bar{w}(L,t) = 0, \qquad \tfrac{\pd \beta}{\pd x}(0,t) = 0,\quad  \tfrac{\pd \beta}{\pd x}(L,t) = 0 
\end{equation}
As a result, the nominal model admits lower bending stiffness near the boundary, causing small but systematic deviations in mode shapes, resonance frequencies, and phase relations compared with the true response.

\paragraph{Discrepancy modeling}
Within the proposed GPLFM framework, the effect of the omitted rotary spring is treated as a localized MFE which is represented by latent forces in the reduced modal domain (four modes; see \cref{sec:DOE}). Despite the local nature of this discrepancy in the physical domain, the use of global nominal mode shapes for projection still enables the GP-based latent force model to reconcile the mismatch effectively during inference, providing accurate displacement estimation and reasonable predictive rectification.

\paragraph{Results} 
\cref{fig:all_examples_est} (third row) compares the spatio–temporal displacement fields of the true and nominal systems with the GPLFM-estimated field over the representative window 3.5–3.6 \si{s}. The GP-augmented nominal model recovers the true displacement field nearly exactly under noiseless data, and preserves the dominant deformation pattern even at 5\% measurement noise. Quantitative errors reported in \cref{tab:NMSE_for_example_1_to_4} (third row) indicate a reduction in NMSE exceeding 99\% relative to the nominal baseline for the estimation stage.

\begin{figure}[H]
    \centering
    \includegraphics[width=0.85\linewidth]{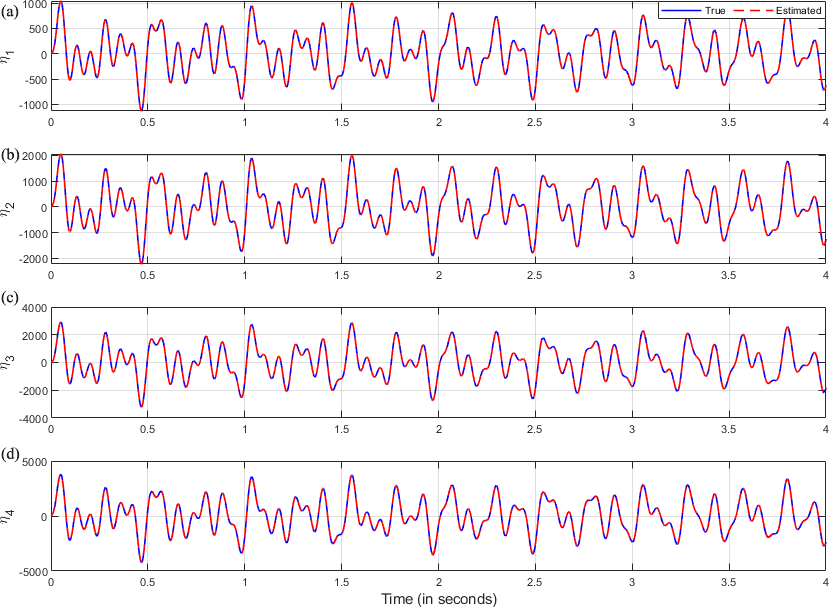}
     \caption{Example 3 (Misspecified boundary condition): Comparison of inferred modal latent forces (red dashed) and ground-truth modal discrepancies (blue solid) for the four retained modes under 0\% noise. The GPLFM accurately recovers all modal latent forces, effectively capturing the localized boundary effect introduced by the rotary stiffness at the beam end despite using global nominal modes for projection.}
    \label{fig:example3_discrepancy_est_noise0}
\end{figure}

\begin{figure}[H]
    \centering
    \includegraphics[width=0.85\linewidth]{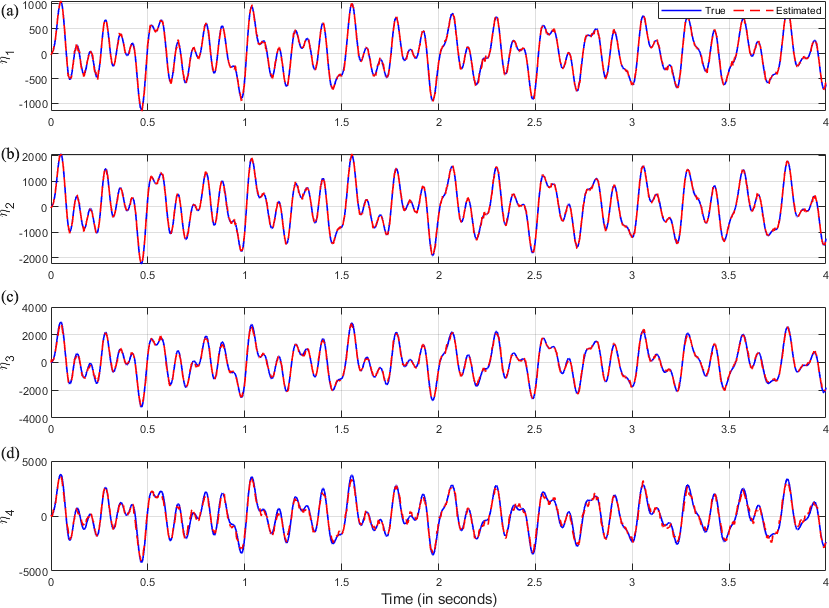}
    \caption{Example 3 (Misspecified boundary condition): Inferred modal latent forces (red dashed) versus ground-truth modal discrepancies (blue solid) for 1\% measurement noise. The dominant lower modes (1 and 2) remain well captured, whereas higher modes (3 and 4) show near-zero estimation as their small response contributions fall below the noise floor.}
    \label{fig:example3_discrepancy_est_noise1}
\end{figure}

\begin{figure}[H]
    \centering
    \includegraphics[width=0.85\linewidth]{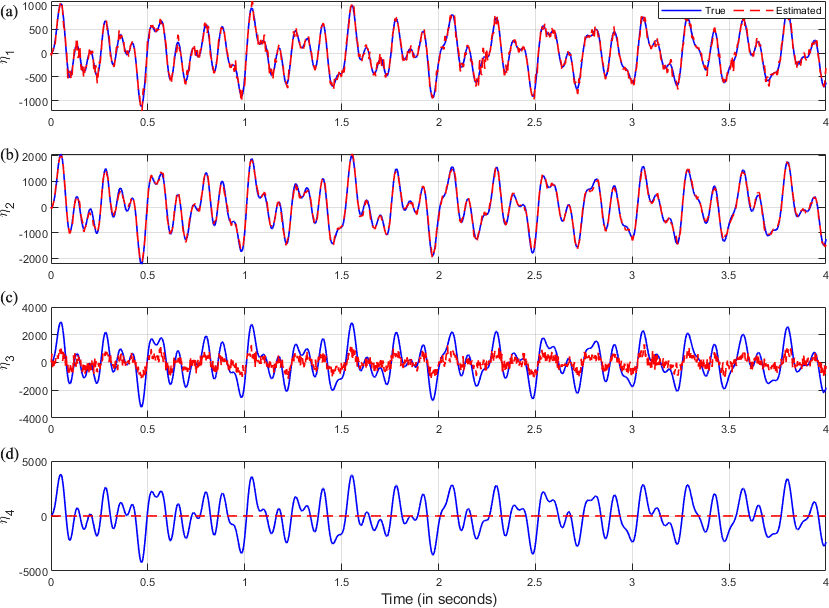}
    \caption{Example 3 (Misspecified boundary condition): Inferred modal latent forces (red dashed) versus ground-truth modal discrepancies (blue solid) for 5\% measurement noise. The GPLFM continues to capture the first modal latent force that encodes the boundary-induced stiffness effects, while higher-mode discrepancies (modes 2–4) diminish to near zero due to their negligible dynamic energy and unidentifiability below the noise threshold.}
    \label{fig:example3_discrepancy_est_noise5}
\end{figure}

\cref{fig:example3_discrepancy_est_noise0,fig:example3_discrepancy_est_noise1,fig:example3_discrepancy_est_noise5} present the inferred modal latent forces compared with the ground-truth discrepancies for 0\%, 1\%, and 5\% noise, respectively. Under noiseless conditions, the latent-force inference captures all four modal latent forces accurately, hence addressing the localized boundary effect introduced by the rotary stiffness. As noise increases, the inferred higher-mode discrepancies (particularly modes~3 and~4) gradually decay toward zero, as their modal responses contribute negligibly to the overall deformation and fall below the noise floor. Nonetheless, the dominant lower modes---especially the first two---remain robustly reconstructed, allowing reliable recovery of the overall dynamics.

\cref{fig:all_examples_pred} (third row) shows the rectified model predictions under unseen test excitation (see \cref{sec:DOE}). The rectified model consistently outperforms the nominal baseline, correcting both phase and amplitude distortions caused by the boundary rotary stiffness omission. While accuracy naturally decreases with larger noise levels due to degraded latent-force estimation and NN fitting, \cref{tab:NMSE_for_example_1_to_4} confirms substantial NMSE reduction even at 5\% noise, demonstrating the framework's robustness in handling spatially localized MFEs.


\subsection{Example 4: Misspecified constitutive relation}
This example investigates MFEs that arise due to discrepancies in the assumed material constitutive behavior. While the previous cases focused on kinematic and boundary-condition mismatches, this case examines whether the proposed GPLFM framework can address errors stemming from incorrect material modeling---a key challenge in digital-twin calibration and reduced-order modeling of nonlinear structures. The \emph{true} beam follows Timoshenko dynamics with a Neo-Hookean hyperelastic constitutive relation, whereas the \emph{nominal} beam is assumed to be linear elastic. To activate material nonlinearity, the excitation amplitudes are increased by an order of magnitude compared with the previous examples (see \cref{sec:DOE}), thereby driving the true beam into a nonlinear deformation regime that deviates from the nominal linear model.


\paragraph{True model}
A compressible Neo-Hookean strain energy function describes the hyperelastic behavior of the true Timoshenko beam,
\begin{align}
    W(I_1, J) = \frac{E}{4(1+\nu)}(J^{-2/3}I_1 - 3) + \frac{E}{6(1-2\nu)}(J - 1)^2,
\end{align}
where $E$ is Young’s modulus, $\nu$ is Poisson’s ratio, $I_1 = \mathrm{tr}(\matr{C})$ is the first invariant of the right Cauchy-Green deformation tensor $\matr{C} = \matr{F}^\top \matr{F}$, and $J = \det(\matr{F})$ is the Jacobian of deformation. The first Piola-Kirchhoff stress is obtained from
\begin{align}
    \matr{P} = \frac{\partial W}{\partial \matr{F}},
\end{align}
with $\matr{F}$ denoting the deformation gradient. The true dynamic response is computed using finite-element simulations in \textsc{Abaqus 2018} \cite{abaqus2018}, ensuring the inclusion of geometric and material nonlinearities.

\paragraph{Nominal model}
The nominal beam adopts a linear elastic isotropic constitutive law, valid under small strains, with strain energy density
\begin{align}
    W = \frac{1}{2}\lambda (\mathrm{tr}\,\boldsymbol{\epsilon})^2 + \mu\, \boldsymbol{\epsilon} : \boldsymbol{\epsilon},
\end{align}
where $\boldsymbol{\epsilon}$ is the infinitesimal strain tensor, and $\lambda$ and $\mu$ are the Lamé parameters related to $(E,\nu)$ by the standard isotropic relations. The Cauchy stress follows $\boldsymbol{\sigma} = \frac{\partial W}{ \partial \boldsymbol{\epsilon}}$. This linear approximation is computationally efficient but fails to capture the nonlinear stress–strain coupling present in the true hyperelastic system.


\paragraph{Discrepancy modeling}
Within the GPLFM framework, the MFEs resulting from the misspecified constitutive relation are modeled as latent forces in the reduced modal domain (four modes; see \cref{sec:DOE}). These latent forces encode the unmodeled nonlinear stress contributions arising from the hyperelastic constitutive behavior, enabling the nominal linear model to reconcile with the true nonlinear response during training and to generalize to unseen excitations.

\paragraph{Results} 
\cref{fig:all_examples_est} (fourth row) compares the spatio–temporal displacement fields of the true (hyperelastic) beam, the nominal linear model, and the GP-augmented nominal estimates for the representative time window 3.5–3.6 \si{s}. The GPLFM-corrected model reproduces the nonlinear displacement field nearly exactly under noiseless data and captures the primary response features even with 5\% noise. Quantitative results summarized in \cref{tab:NMSE_for_example_1_to_4} (fourth row) show that the augmented model achieves over 96\% reduction in NMSE relative to the nominal baseline, despite the strong material nonlinearity.

\begin{figure}[H]
    \centering
    \includegraphics[width=0.75\linewidth]{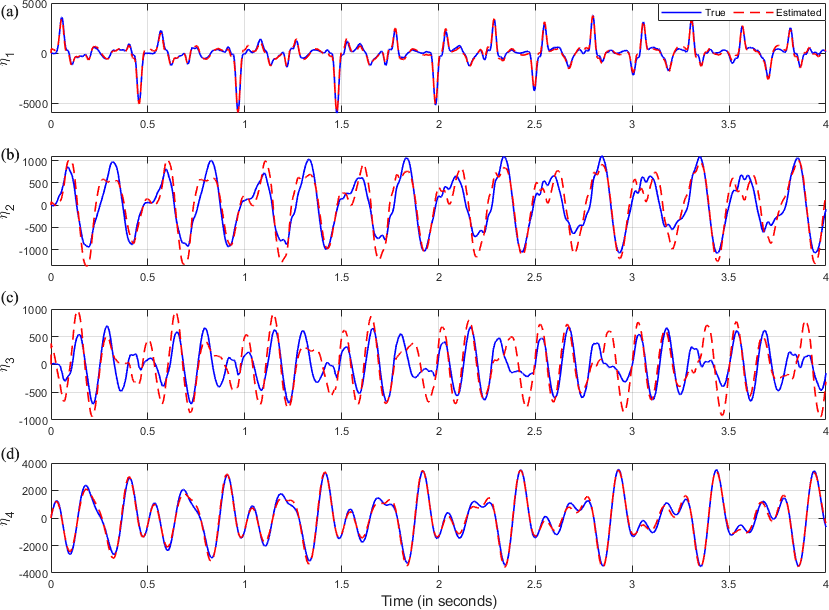}
    \caption{Example~4 (Misspecified constitutive relation): Comparison of inferred modal latent forces (red dashed) and ground-truth nonlinear discrepancies (blue solid) for the four retained modes under 0\% noise. The GPLFM accurately reconstructs all modal latent forces, capturing the nonlinear stress–strain effects introduced by the true Neo-Hookean material while the nominal model remains linear elastic.}
    \label{fig:example4_discrepancy_est_noise0}
\end{figure}

\begin{figure}[H]
    \centering
    \includegraphics[width=0.75\linewidth]{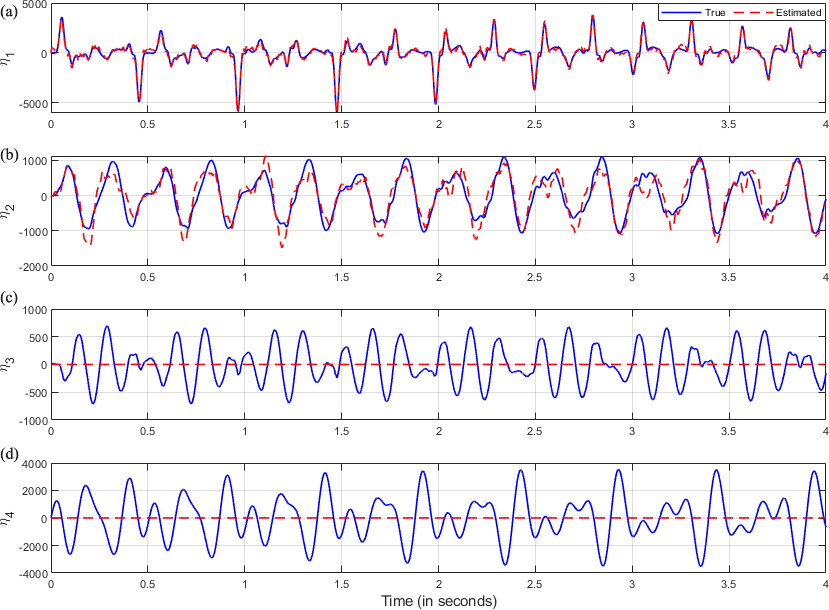}
    \caption{Example~4 (Misspecified constitutive relation): Inferred modal latent forces (red dashed) versus ground-truth nonlinear discrepancies (blue solid) for 1\% measurement noise.
    The dominant modal latent forces remain well identified, faithfully reproducing the primary nonlinear material effects, whereas higher-order components show mild attenuation with the onset of noise.}
    \label{fig:example4_discrepancy_est_noise1}
\end{figure}

\begin{figure}[H]
    \centering
    \includegraphics[width=0.75\linewidth]{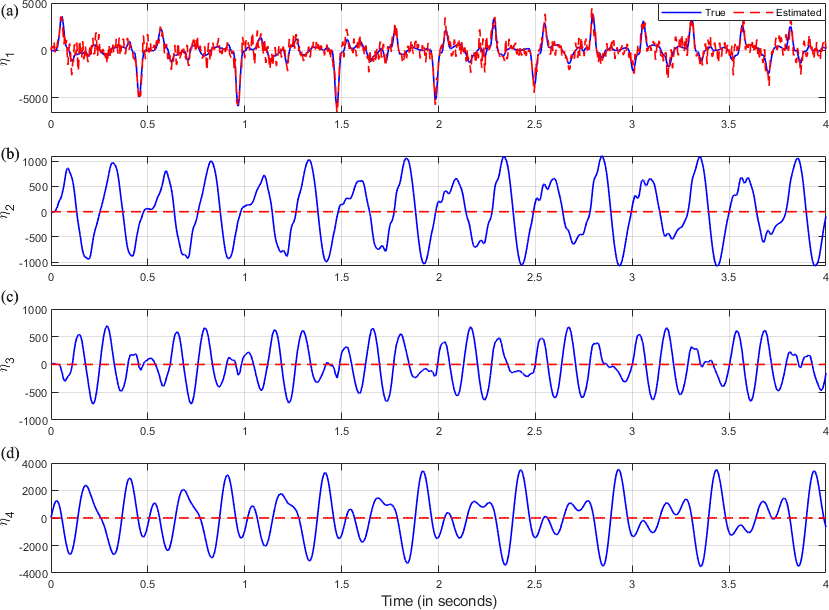}
    \caption{Example~4 (Misspecified constitutive relation): Inferred modal latent forces (red dashed) versus ground-truth nonlinear discrepancies (blue solid) for 5\% measurement noise.
    The GPLFM continues to capture the principal low-frequency latent force representing the missing nonlinear constitutive effects, while higher-mode contributions (modes 3–4) decay toward zero as their signal energy falls below the noise threshold.}
    \label{fig:example4_discrepancy_est_noise5}
\end{figure}

\cref{fig:example4_discrepancy_est_noise0,fig:example4_discrepancy_est_noise1,fig:example4_discrepancy_est_noise5} present the inferred modal latent forces versus the true modal discrepancies for 0\%, 1\%, and 5\% measurement noise, respectively. Under noiseless conditions, the latent-force inference captures all four modal discrepancies, effectively representing the missing nonlinear material response. With increasing noise, the contribution of higher modes (particularly modes~3 and~4) falls below the noise floor, leading to diminished amplitude in their inferred latent forces, while the dominant lower modes remain reasonably reconstructed.

\cref{fig:all_examples_pred} (fourth row) illustrates predictive performance under unseen test excitations. The rectified model, enriched by the inferred latent forces, outperforms the nominal linear model in both amplitude and phase with $0\%$ noise, even though the non-linear material behaviour is entirely absent from the nominal formulation. For the case of $5\%$ noise the predictive performance is poor when looked at a particular small time window 3.5-3.6 \si{s} but over the full time period 0-4 \si{s} \cref{tab:NMSE_for_example_1_to_4} confirms that the GPLFM is still able to achieve decent NMSE reductions, demonstrating its usefulness in compensating for complex constitutive mismatches.


\subsection{Example 5: Localized stiffness loss in a 120-DOF IASC–ASCE benchmark building}
\begin{figure}[H]
    \centering
    \begin{subfigure}[b]{0.3\textwidth}
        \centering
        \includegraphics[width=\textwidth]{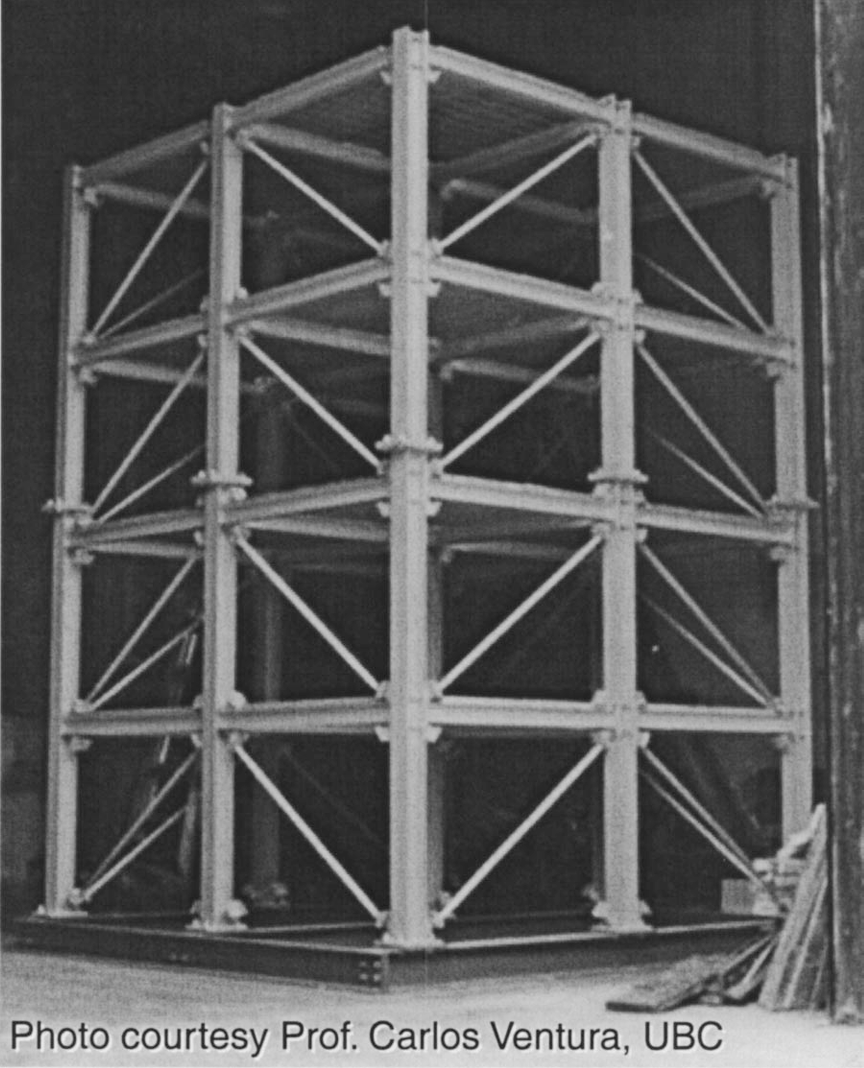}
        \caption{}
        \label{fig:sub1}
    \end{subfigure}
    \hspace{1em}
    \begin{subfigure}[b]{0.45\textwidth}
        \centering
        \includegraphics[width=\textwidth]{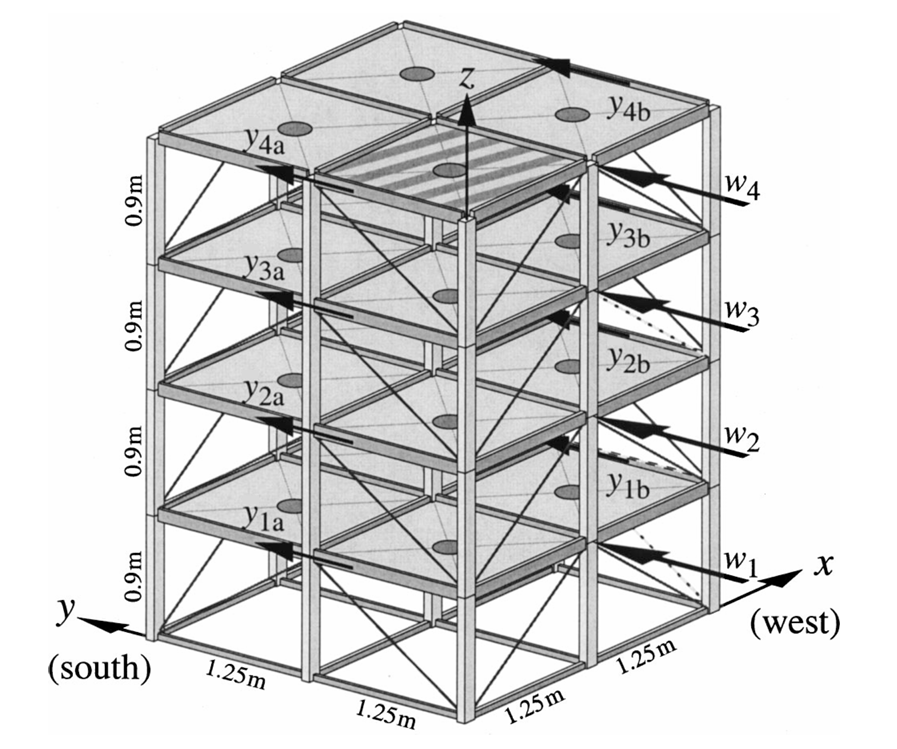}
        \caption{}
        \label{fig:sub2}
    \end{subfigure}
    \caption{Example~5 (Localized stiffness loss). (a) Experimental four-story IASC–ASCE benchmark structure \cite{johnson2004phase} and (b) its numerical 120-DOF simulation model (adapted from \cite{johnson2004phase}). Excitations $w_i(t)$ are applied in the $y$-direction at all four floors, while displacement measurements $y_{ij}(t)$ are collected at four sensor locations per floor---two in each horizontal direction. For clarity, the $x$-direction measurements ($y_{ic}$ and $y_{id}$) are omitted from the schematic.}
    \label{fig:120DOF}
\end{figure}

This final case study evaluates the proposed framework on the well-known 120-DOF shear building from the IASC–ASCE benchmark suite \cite{johnson2004phase}, representing a realistic high-dimensional structural system. 
The model, shown in \cref{fig:120DOF}, idealizes each floor as a rigid slab, such that the horizontal translations in two orthogonal directions and in-plane rotation are identical at all points within each floor. The remaining three degrees of freedom: vertical translation and two out-of-plane rotations are unconstrained for each floor. The structure behaves linearly within the excitation range considered, but a local loss of stiffness is introduced to emulate damage.

\paragraph{Model description}
The true system follows the ``Pattern 1'' damage configuration of the IASC--ASCE benchmark structure, wherein the bracing members of the first story are removed to simulate a localized stiffness loss. 
The nominal model corresponds to the undamaged configuration of the same building. 
The structure is excited in the $y$-direction at all four floor levels using distinct loading time histories for training and testing. 
Each excitation is a narrow-band Gaussian process obtained by filtering white Gaussian noise through a fourth-order Butterworth low-pass filter with a passband of $[0, 30]~\si{Hz}$. 
A reduced-order representation of the 120-DOF model retaining the first nine vibration modes identified from the averaged response spectrum was found to adequately capture the essential dynamic behavior while maintaining computational efficiency.

Table \ref{tab:example_5_natural_frequency} lists the natural frequencies of the undamaged and damaged structures for the first nine retained modes.
The table also identifies each mode's dominant motion direction ($x$, $y$, or $\theta$ about $z$).

\paragraph{Measurement setup}
Following the benchmark specification, sixteen sensors are distributed uniformly across the four floors (four per floor, two in each horizontal direction), as shown in \cref{fig:120DOF}. 
Displacement measurements from these locations are employed here for GPLFM inference. Preliminary analyses indicated that using acceleration-only measurements provided poor estimation of modal latent forces, particularly under noise, which in turn deteriorated the rectified predictions under unseen excitations. Displacement data, in contrast, provided more accurate recovery of both modal states and latent forces, and are therefore used throughout this study. 

\paragraph{Noise level}
Measurement noise is modeled as additive white Gaussian noise of 1\%, consistent with the earlier examples. 
For higher noise levels (e.g., 5\%), the latent-force estimation---particularly with using Mat\'ern-$\tfrac{1}{2}$ kernel---becomes noisy and unreliable, leading to poor rectification; therefore, only the 1\% noise case is presented.

\begin{figure}[H]
    \centering
    \includegraphics[width=0.90\linewidth]{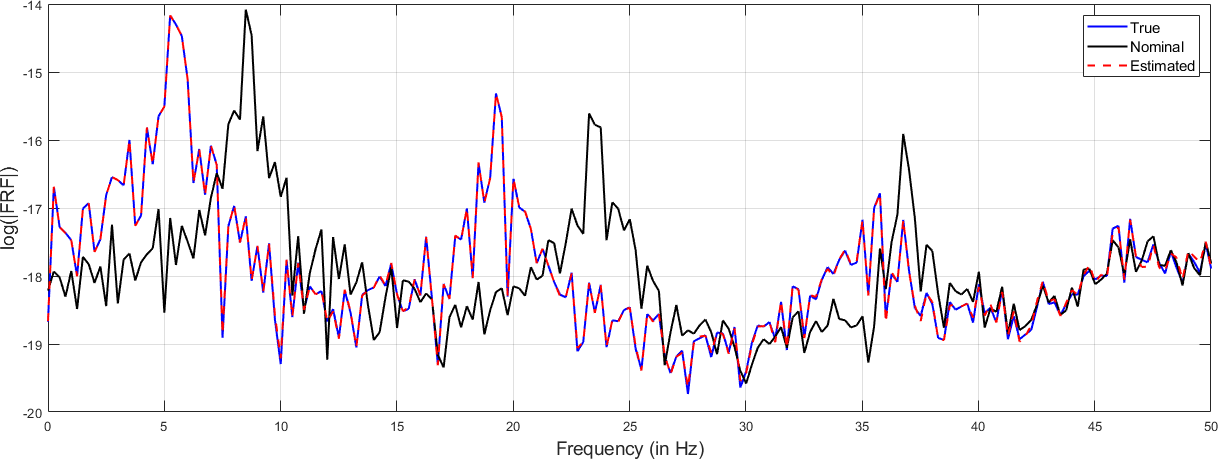}
    \caption{Example 5 (Localized stiffness loss) Mean logarithm of frequency-response functions (FRFs) computed by averaging the magnitude spectra of all 120 DOFs, including translational ($x$, $y$) and rotational ($\theta$) components. The estimated FRF reproduces the resonance peaks of the true damaged structure, with dominant peaks corresponding to the $y$-directional modes (1, 4, 6, and 9) highlighted in Table \ref{tab:example_5_natural_frequency}.}

    \label{fig:example5_FRF_est}
\end{figure}

\begin{figure}[H]
    \centering
    \includegraphics[width=0.75\linewidth]{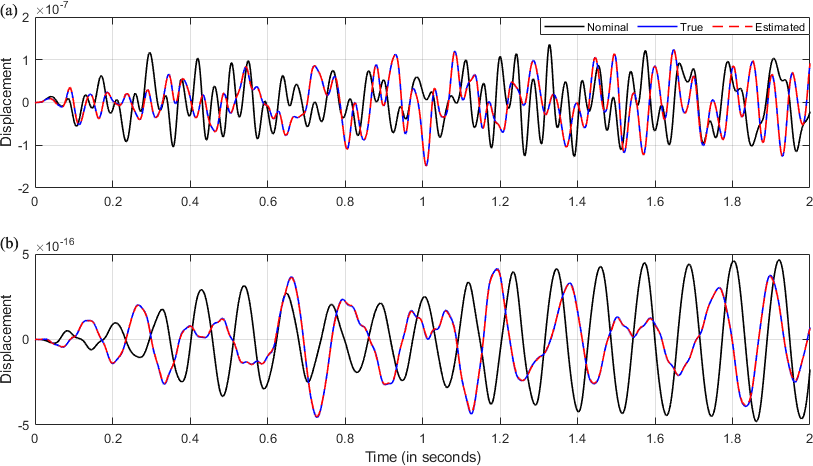}
    \caption{Example 5 (Localized stiffness loss) Comparison of estimated and true displacement responses at two representative third-floor locations, one in the $x$-direction and one in the $y$-direction—under 1\% noise.}
    \label{fig:example5_disp_est_noise1}
\end{figure}

\paragraph{Results}
Because the excitation is applied exclusively in the $y$-direction, these $y$-dominant modes contribute most strongly to the measured responses and, as shown below, yield the most accurate latent-force estimates.
The GPLFM-based inference reconstructs displacement responses accurately at both instrumented and uninstrumented locations; representative comparisons on the third floor in the $x$- and $y$-directions are shown in \cref{fig:example5_disp_est_noise1}. 
For spectral validation, \cref{fig:example5_FRF_est} presents the \emph{mean} of the logarithm of frequency response function (FRF), obtained by averaging the log-magnitude spectra of all 120 DOFs, including translational ($x$, $y$, $z$) and rotational ($\theta$) components. The estimated FRF closely follows that of the true damaged system across the 0-50 Hz range encompassing the nine retained modes, with prominent peaks corresponding to the $y$-directional modes identified in Table \ref{tab:example_5_natural_frequency}.

\begin{figure}[H]
    \centering
    \includegraphics[width=0.9\linewidth]{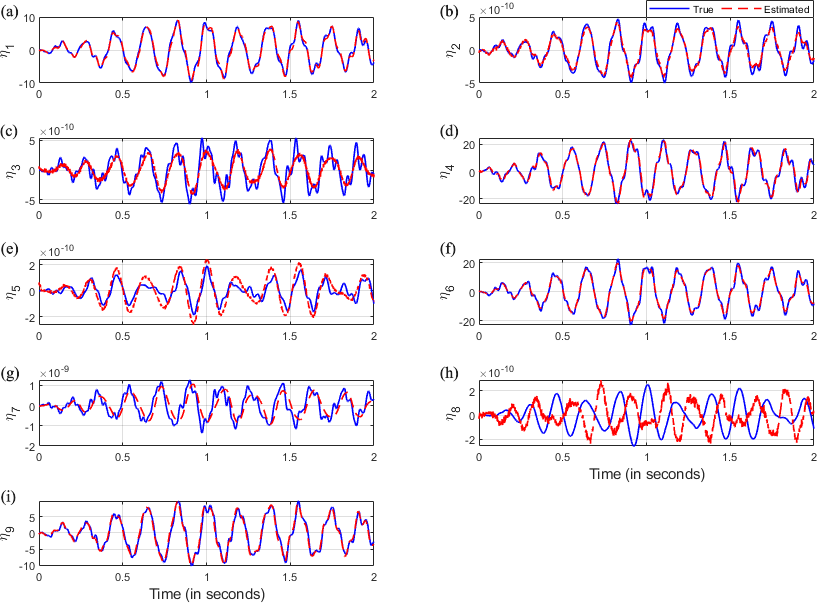}
    \caption{Example 5 (Localized stiffness loss) Inferred modal latent forces (red dashed) versus ground-truth discrepancies (blue solid) for the first nine retained modes under 1\% noise. Accurate estimation is achieved for the strongly excited $y$-directional modes (1, 4, 6, and 9) and the coupled rotational mode 7.}
    \label{fig:example5_discrepancy_est_noise1}
\end{figure}

The inferred modal latent forces, shown in \cref{fig:example5_discrepancy_est_noise1}, agree closely with the ground-truth discrepancies in these dominant $y$-modes (1, 4, 6, and 9) and in the $\theta$-mode 7, which couples strongly with $y$-motion. Modes with weaker excitation energy exhibit comparatively smaller latent-force magnitudes, consistent with their lower modal participation in the overall response.

\begin{figure}[H]
    \centering
    \includegraphics[width=0.75\linewidth]{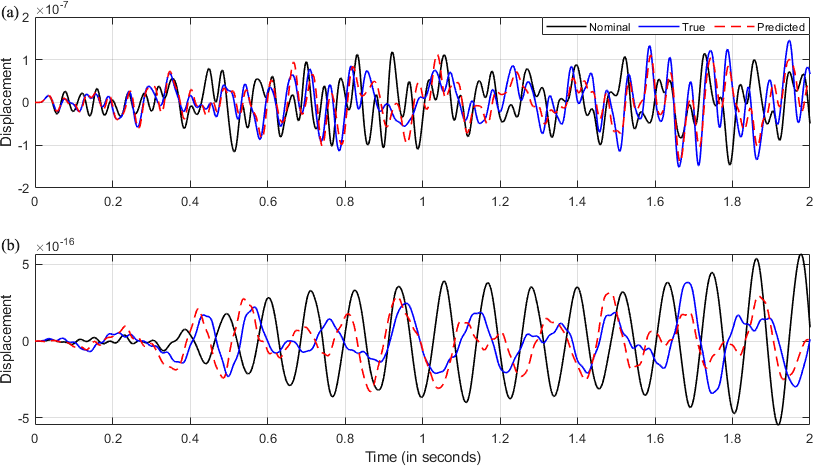}
    \caption{Example 5 (Localized stiffness loss) Comparison of predicted and true displacement responses at two representative third-floor locations under unseen excitations.}
    \label{fig:example5_state_pred_noise1}
\end{figure}

Finally, embedding the trained neural surrogate into the nominal model yields the rectified predictive simulator. Its displacement predictions under unseen excitations---different realizations of filtered white Gaussian noise---are shown in \cref{fig:example5_state_pred_noise1}. 
The rectified model achieves NMSE reductions of approximately 85–86\% relative to the nominal baseline at the same locations on the third floor, demonstrating decent prediction of the true dynamic behavior of the damaged system. 

\begin{table}[H]
\centering
\caption{First nine natural frequencies (Hz) of the undamaged and damaged benchmark structures, with each mode labeled by its dominant motion direction ($x$, $y$, or $\theta$).}
\label{tab:example_5_natural_frequency}
\begin{tabular}{|c|c|c|c|c|c|c|c|c|c|}
\hline
\textbf{Model} & Mode 1 & Mode 2 &Mode 3 &Mode 4 & Mode 5 &Mode 6 &Mode 7 &Mode 8 &Mode 9  \\
 \hline
 \textit{Undamaged} & 8.59 $y$ & 9.18 $x$ & 14.58 $\theta$ & 23.45 $y$ & 25.95 $x$ & 36.81 $y$ & 40.65 $\theta$ & 42.21 $x$ &46.98 $y$  \\
 \hline
 \textit{Damaged} & 5.47 $y$ & 7.37 $x$ & 9.69 $\theta$ & 19.31 $y$ & 22.77 $x$ & 34.18 $y$ & 35.18 $\theta$ & 40.66 $x$ & 46.73 $y$ \\
 \hline
\end{tabular}
\end{table}

\section{Discussion} \label{sec:discussion}
The proposed framework presents an approach for diagnosing and rectifying MFEs in high-dimensional structural dynamical systems.
By integrating GPLFMs with a neural mapping surrogate, it systematically captures the latent corrective dynamics that reconcile the nominal physics-based models with measured system responses.

A key feature of the proposed approach is its \emph{modal-space formulation}, which enables inference and prediction in a reduced-order setting. This reduction provides three main benefits:
\begin{itemize}
    \item \textit{Dimensionality reduction}: Inference is performed in a low-dimensional space governed by a few dominant modes rather than the full FE structural model, substantially lowering computational costs.

    \item \textit{Fast rectified prediction}: Once trained, the rectified model operates in the same reduced modal domain, making rapid predictions under unseen excitations.

    \item \textit{Mesh invariance}: Because both training and inference occur entirely in modal space, the approach remains independent of FE mesh resolution. This mesh-invariance allows the trained surrogate to be transferable across differing discretizations or spatial refinements---an important feature for digital twins that evolve over multiple fidelity levels.
\end{itemize}

Across all numerical case studies, the approach consistently inferred meaningful modal latent forces and delivered good rectified predictions under unseen loading conditions, illustrating its scalability for high-dimensional structures. 

It is worthwhile to note that the external excitation was assumed to be perfectly known and noise-free. If the excitation were contaminated by noise or only partially characterized, the inferred latent forces---which only assume state dependence of the MFEs in this study---would inevitably absorb errors in excitation, leading to incorrect rectification. Future extensions may integrate stochastic excitation models or Bayesian identification schemes to jointly infer both input uncertainty and model bias, as explored in recent studies such as \cite{lourens2025bias}.

All examples in this work employed displacement measurements, which provided better modal latent-force inference compared to acceleration-only data. We found that acceleration measurements produce high-frequency noise in the modal latent forces and lead to degraded GPLFM performance, especially in estimating higher-mode latent forces. Displacements (or strain-type) measurements are smoother and more conducive to reliable latent-force recovery.

The GPLFM formulation offers considerable flexibility in modeling the temporal characteristics of MFEs through kernel choice \cite{vettori2023assessment}. While a Mat\'ern-$\tfrac{1}{2}$ kernel was employed throughout this study for its simplicity and computational tractability, this choice is not universally optimal. Kashyap et al. \cite{kashyap2024gaussian} has shown that higher-order kernels (e.g., Mat\'ern–$\tfrac{3}{2}$ or Mat\'ern-$\tfrac{5}{2}$) can yield more accurate latent-force estimates under high noise or more complex temporal dynamics, but they nearly double the state-space dimension and associated computational cost.
The Mat\'ern-$\tfrac{1}{2}$ kernel thus represents a balanced compromise between accuracy and efficiency.
The GPLFM framework, however, remains general and can incorporate any stationary kernel admitting an SDE representation, enabling customization based on the temporal structure of model discrepancies.

A few additional aspects merit discussion. First, the framework's inference performance remains sensitive to measurement noise, particularly when the latent dynamics are weak or poorly excited.
Second, the present implementation uses independent latent-force GPs with block-diagonal process covariances for tractability; correlated GP formulations (capturing dependencies between each modal latent force and its corresponding modal state) could further improve identifiability across modes, albeit at higher computational cost during inference.

Third, the method assumes a known nominal modal basis derived from deterministic system matrices.
In problems involving parameter uncertainty or system identification where mass or stiffness matrices themselves are uncertain, a single modal basis may be inadequate.
Extending the approach to handle families of modal bases derived from uncertain parameter realizations represents a natural next step.

Fourth, the current mapping between modal states and modal latent forces is deterministic and instantaneous; temporal history-dependence of MFEs is not yet modeled. Future extensions could explore recurrent or convolutional network architectures to capture such dynamic dependencies.
Introducing Bayesian or variational formulations could enable explicit \textit{uncertainty quantification} in both latent-force estimation and predictive inference, enhancing the framework's reliability for decision-making in digital-twin applications. 

Finally, while this work focused on nominally linear models, the methodology can be extended to nonlinear dynamics using nonlinear filters---such as the Unscented Kalman filter or Ensemble Kalman filter---in the inference stage.

\section{Conclusions}\label{sec:conclusion}
This study presented a framework for diagnosing and rectifying model-form errors (MFEs) in structural dynamical systems by integrating Gaussian Process Latent Force Models (GPLFMs) with a neural mapping surrogate. The framework performs joint probabilistic inference of modal states and latent discrepancies within a reduced modal domain, enabling predictive rectification under unseen excitations.

The proposed approach offers the following contributions. 
First, it formulates a \emph{hybrid physics-consistent latent-force modeling framework}, wherein the nominal governing equations provide the physical backbone while data-driven GPs infer the latent dynamics required for model rectification. 
Second, by operating entirely in modal coordinates, the framework achieves substantial dimensionality reduction while remaining independent of the finite element (FE) mesh, thereby ensuring \emph{mesh invariance} and facilitating transferability across models with varying spatial discretizations or fidelity levels. 
Third, the framework's versatility was demonstrated through a diverse suite of five representative examples encompassing modeling simplification (Euler–Bernoulli vs.\ Timoshenko beam), damping mis-specification, boundary condition errors, nonlinear constitutive behavior, and localized stiffness loss in a 120-DOF IASC–ASCE benchmark building. 
Across these diverse cases, the methodology consistently inferred latent forces and produced accurate rectified predictions under unseen loading conditions, underscoring both its \emph{scalability}. 

Future research may focus on enhancing robustness under noisy or uncertain measurements, introducing correlated GPs to capture state-latent-force dependencies, and embedding probabilistic formulations for explicit uncertainty propagation in prediction. Extending the approach to nominally nonlinear systems through nonlinear filters such as the Unscented or Ensemble Kalman filter represents another promising direction toward reliable, real-time digital-twin applications.

\newpage

\appendix
\begin{appendices}
\section{Conversion of Gaussian process into State-Space Model (SSM)} \label{appendix: GP to SSM}
In this work, model discrepancy is modeled as a zero-mean Gaussian process (GP) with a stationary covariance function. To enable integration with the state-space dynamics of the nominal system, we convert the GP representation into an equivalent linear time-invariant (LTI) state-space model (SSM), leveraging the fact that certain classes of GP kernels admit such formulations~\cite{sarkka2013spatiotemporal}.

Consider a scalar GP \( y(t) \sim \mathcal{GP}(0, C(\tau)) \), where \( C(\tau) \) is the covariance function parameterized by time lag \( \tau \). For covariance functions that correspond to the impulse response of linear stochastic differential equations (SDEs), the GP can be equivalently described as the solution to a stochastic differential equation of the form:

\begin{equation}
a_n \frac{d^n y(t)}{dt^n} + \cdots + a_1 \frac{dy(t)}{dt} + a_0 y(t) = w(t)
\label{eq:sde}
\end{equation}

Here, \( w(t) \) denotes a Gaussian white noise process with constant power spectral density \( q_c \), and \( a_i \) are real-valued coefficients that define the dynamics of the latent process.

Taking the Fourier transform of  \cref{eq:sde} leads to a transfer function formulation:

\[
F(i\omega) = \frac{G(i\omega)}{W(i\omega)} = \frac{1}{a_n(i\omega)^n + \cdots + a_0}
\]

This leads to the power spectral density:

\[
S(\omega) = |G(i \omega)|^2 q_c
\]

In accordance with the Wiener–Khinchin theorem, the covariance function \( C(\tau) \) of a stationary GP can be obtained by taking the inverse Fourier transform of \( S(\omega) \):

\[
C(\tau) = \frac{1}{2\pi} \int_{-\infty}^\infty S(\omega)\, e^{i \omega \tau}\, d\omega
\]

To illustrate the process, consider an Mat\'ern-$\tfrac{1}{2}$ (exponential) kernel:

\[
C(\tau) = \alpha^2 e^{-|\tau| / l}
\]

The Fourier transform of this kernel yields:

\[
S(\omega) = \frac{2 \alpha^2 l}{1 + (l \omega)^2}
\]

This spectral density corresponds to the solution of a first-order SDE:

\[
\frac{dy(t)}{dt} = -\frac{1}{l} y(t) + w(t)
\]

which represents a stable Ornstein–Uhlenbeck process. Its equivalent state-space model formulation is:

\begin{align}
\dot{\xi}(t) &= -\frac{1}{l} \xi(t) + w(t) \\
\eta(t) &= \xi(t)
\end{align}

Here, \( \xi(t) \) is the state of the discrepancy process and \( \eta(t) \) is its output, which is added to the dynamics of the physical system model (as formulated in \cref{sec:met_3}). This approach preserves dynamic consistency and ensures asymptotic stability of the latent input model.

For higher-order kernels such as Mat\'ern or squared exponential (SE) functions, similar derivations can be made under the Gaussian process–SSM duality, provided their spectral densities admit rational approximations~\cite{sarkka2013spatiotemporal}.

\section{MAP optimization of GP hyperparameters}\label{appendix:MAP optimization of GP hyperparameters}
In this framework, each latent force $\eta_i$ is described as an independent Gaussian process (GP), with its behavior characterized by a small set of hyperparameters $\ve{\theta}_i = [\alpha_i, \ell_i]$. The term $\alpha_i$ typically serves as a scaling factor or amplitude parameter, while $\ell_i$ represents the characteristic length-scale governing smoothness and correlation structure of the process. When an $m$ modes is considered, assigning one latent force to each modes naturally leads to $m$ latent GPs being defined. As a result, the complete parameter vector can be expressed as $\ve{\theta} = [\ve{\theta}_1, \ldots, \ve{\theta}_n]$, giving a total of $2m$ free parameters.  

From a Bayesian standpoint, this parameter vector $\ve{\theta}$ is modeled as a random variable rather than as a fixed unknown constant. Consequently, the inference task requires computing the posterior density of the hyperparameters given the observations. Making use of Bayes’ rule, the posterior is given by  
\begin{equation*}
     p \left(\boldsymbol{\theta} \mid \mathbf{y}_{1:N_t} \right) 
    \propto p \left(\mathbf{y}_{1:N_t} \mid \boldsymbol{\theta} \right) \, p(\boldsymbol{\theta}),   
\end{equation*}

where the likelihood term $p(\mathbf{y}_{1:N} \mid \boldsymbol{\theta})$ encodes how well the parameters explain the experimental data, and the prior $p(\boldsymbol{\theta})$ regularizes the inference by embedding prior knowledge about plausible parameter ranges.  

In state-space formulations, the available data typically consists not only of the observed outputs or measurements $\mathbf{y}_{1:N}$, but also of the hidden state sequence $\mathbf{x}_{1:N}$. Since the latent states are not observed directly, they are integrated out in order to derive the marginal posterior distribution over the hyperparameters. This marginalization step introduces a significant computational bottleneck, as the evaluation of the marginal likelihood $p(\mathbf{y}_{1:N} \mid \boldsymbol{\theta})$ is analytically intractable in most practical cases. 

To address this, appropriate choices of prior distributions are imposed on the hyperparameters. In particular, independent Student-$t$ distributions are adopted in this work. Using the generic parameterization $\mathcal{T}(\mu, v, \nu)$, where $\mu$ is the mean, $v$ the variance, and $\nu$ the degrees-of-freedom, the log-prior distribution takes the form  
\begin{equation*}
     \log p(\boldsymbol{\theta}) = \sum_{j=1}^m \Big( \log \mathcal{T} (\alpha_j; 10^4, 10^2, 1) + \log \mathcal{T} (\ell_j; 0.1, 10^{-2}, 1) \Big)   
\end{equation*}  

The marginal likelihood itself admits a recursive factorization in sequential data models, given as  
\begin{equation*}
    p \left( \mathbf{y}_{1:N_t} \mid \boldsymbol{\theta} \right) 
    = \prod_{k=1}^{N_t} p \left( \ve{y}_k \mid \mathbf{y}_{1:k-1}, \boldsymbol{\theta} \right).    
\end{equation*}

For linear Gaussian state-space models, this quantity can be obtained efficiently from the Kalman filter innovations $\ve{e}_k$ and their covariance matrices $\matr{S}_k$. Explicitly, the log marginal likelihood can be expressed as  
\begin{equation*}
     \log p \left( \mathbf{y}_{1:N_t} \mid \boldsymbol{\theta} \right)
    = \sum_{k=1}^{N_t} \log p \left( \ve{y}_k \mid \ve{y}_{1:k-1}, \boldsymbol{\theta} \right)
    = -\sum_{k=1}^{N_t} \left( \log \det \matr{S}_k + \ve{e}_k^\top \matr{S}_k^{-1} \ve{e}_k \right).   
\end{equation*}

This recursive computation, facilitated by the Kalman filter, efficiently compresses the likelihood evaluation into a series of innovation terms, avoiding the need to explicitly integrate over the high-dimensional latent state trajectory.  

Once the log-posterior has been constructed by combining the log-likelihood and log-prior terms, the inference problem reduces to identifying the maximum a posteriori (MAP) configuration of the hyperparameters:  
\begin{align*}
     \hat{\boldsymbol{\theta}}_{\mathrm{MAP}} &= \arg\max_{\boldsymbol{\theta}} \, \big(\log p \left( \mathbf{y}_{1:N_t} \mid \boldsymbol{\theta} \right) + \log p(\boldsymbol{\theta})\big)   \\
     & = \arg\min_{\boldsymbol{\theta}}
    \left( -\log p \left( \mathbf{y}_{1:N_t} \mid \boldsymbol{\theta} \right) - \log p(\boldsymbol{\theta}) \right) 
\end{align*}

The optimization is conventionally performed in the log-transformed parameter domain, which ensures that the estimated hyperparameters respect the positivity constraints inherent to amplitude and length-scale parameters. This transformation not only enforces physically meaningful values throughout the iterative search but also improves numerical stability during optimization.  

\end{appendices}

\newpage

\bibliography{MDE_PDE_refs}
\end{document}